\documentclass{article}
\usepackage[english]{babel}
\usepackage[utf8]{inputenc} \usepackage[T1]{fontenc}

\usepackage{csquotes}

\usepackage{PRIMEarxiv}

\usepackage{hyperref}       \usepackage{url}            \usepackage{rotating, tabularx}
\newcolumntype{Y}{>{\centering\arraybackslash}X}

\usepackage{array}
\usepackage{booktabs}  \usepackage{siunitx}   \usepackage{multirow}
\usepackage[
    aboveskip=1pt,
    labelfont=bf,
    labelsep=period,
    singlelinecheck=off]{caption}

\usepackage{float}

\usepackage{amsfonts}       \usepackage{amsmath}
\usepackage{amssymb}
\usepackage{nicefrac}       \usepackage{microtype}      \usepackage{lipsum}
\usepackage{fancyhdr}       \usepackage{graphicx}       \usepackage{xcolor}
\graphicspath{{media/}}

\usepackage[utf8]{inputenc}
\usepackage{textcomp,marvosym}
\usepackage{hyperref}
\hypersetup{hidelinks}

\definecolor{o}{rgb}{1, 0.498, 0.055}
\definecolor{b}{rgb}{0.612, 0.271, 0.0}

\usepackage[nohyperlinks,nolist]{acronym}
\acrodef{AD}[AD]{automatic differentiation}
\acrodef{AI}[AI]{artificial intelligence}
\acrodef{ANN}[ANN]{artificial neural network}
\acrodef{BP}[BP]{back-propagation}
\acrodef{BPTT}[BPTT]{back-propagation through time}
\acrodef{CNN}[CNN]{convolutional neural network}
\acrodef{CPU}[CPU]{central processing unit}
\acrodef{CSNN}[ConvSNN]{convolutional spiking neural network}
\acrodef{DNN}[DNN]{deep neural network}
\acrodef{GPU}[GPU]{graphics processing unit}
\acrodef{KL}[KL]{Kullback-Leibler}
\acrodef{LIF}[LIF]{leaky integrate-and-fire}
\acrodef{MLP}[MLP]{multi-layer Perceptron}
\acrodef{PSP}[PSP]{postsynaptic potential}
\acrodef{PSTH}[PSTH]{peri-stimulus-time histogram}
\acrodef{Randman}[Randman]{random manifolds}
\acrodef{ReLU}[ReLU]{rectified linear unit}
\acrodef{ResNet}[ResNet]{residual network}
\acrodef{RNN}[RNN]{recurrent neural network}
\acrodef{RTRL}[RTRL]{real-time recurrent learning}
\acrodef{stochAD}[stochAD]{stochastic automatic differentiation}
\acrodef{SD}[SD]{surrogate derivative}
\acrodef{SGD}[SGD]{stochastic gradient descent}
\acrodef{SG}[SG]{surrogate gradient}
\acrodef{SHD}[SHD]{Spiking Heidelberg Digits}
\acrodef{SMORMS3}[SMORMS3]{squared mean over root mean squared cubed}
\acrodef{SNN}[SNN]{spiking neural network}
\acrodef{SPM}[SPM]{smoothed probabilistic model}
\acrodef{SRM}[SRM]{spike-response model}
\acrodef{STE}[STE]{straight-through estimator}

\usepackage[
    natbib=true,
    sorting=none,
    giveninits=true,
    doi=true,
    url=false,
    maxcitenames=1,
    maxbibnames=42,
    citestyle=numeric-comp]{biblatex}
\DeclareNameFormat{default}{\nameparts{#1}\usebibmacro{name:family-given}
     {\namepartfamily}
     {\namepartgiveni}
     {\namepartprefix}
     {\namepartsuffix}\usebibmacro{name:andothers}}

\title{\Large{Elucidating the theoretical underpinnings of surrogate gradient learning in spiking neural networks}}

\author{
	Julia Gygax$^{1,2}$ \& Friedemann Zenke$^{1,2}$ \\
  $^1$ Friedrich Miescher Institute for Biomedical Research \\
  $^2$ Faculty of Science, University of Basel\\
  Basel, Switzerland\\
  \texttt{\{firstname.lastname\}@fmi.ch} \\
}

\begin{document}

\maketitle

\begin{abstract}
    Training \aclp{SNN} to approximate universal functions is essential for studying information processing in the brain and for neuromorphic computing.
    Yet the binary nature of spikes poses a challenge for direct gradient-based training. \Aclp{SG} have been empirically successful in circumventing this problem, but their theoretical foundation remains elusive.
    Here, we investigate the relation of \aclp{SG} to two theoretically well-founded approaches.
    On the one hand, we consider \aclp{SPM}, which, due to the lack of support for \acl{AD}, are impractical for training multi-layer \aclp{SNN} but provide derivatives equivalent to \aclp{SG} for single neurons.
    On the other hand, we investigate \acl{stochAD}, which is compatible with discrete randomness but has not yet been used to train \aclp{SNN}.
    We find that the latter gives \aclp{SG} a theoretical basis in stochastic \aclp{SNN}, where the \acl{SD} matches the derivative of the neuronal escape noise function.
    This finding supports the effectiveness of \aclp{SG} in practice and suggests their suitability for stochastic \aclp{SNN}.
    However, \aclp{SG} are generally not gradients of a surrogate loss despite their relation to  \acl{stochAD}.
    Nevertheless, we empirically confirm the effectiveness of \aclp{SG} in stochastic multi-layer \aclp{SNN} and discuss their relation to deterministic networks as a special case.
    Our work gives theoretical support to \aclp{SG} and the choice of a suitable surrogate derivative in stochastic \aclp{SNN}.
\end{abstract}

\section{Introduction}
\label{sec: intro}

Our brains efficiently process information in \acp{SNN} that communicate through short stereotypical electrical pulses called spikes.
\acp{SNN} are an indispensable tool for understanding information processing in the brain and instantiating similar capabilities in silico.
Like conventional \acp{ANN}, \acp{SNN} require training to implement specific functions.
However, \ac{SNN} models, which are commonly simulated in discrete time \citep{eshraghian_training_2023}, are not differentiable due to the binary nature of the spike, which precludes the use of standard gradient-based training techniques based on \ac{BP}~\citep{rumelhart_learning_1986}.
There are several ways to overcome this problem.
One can dispense with hidden layers altogether \citep{gutig_tempotron_2006, memmesheimer_learning_2014}, but this limits the network's expressivity and precludes solving more complex tasks.
Alternatively, one makes the neuron model differentiable \citep{huh_gradient_2018} or considers the timing of existing spikes for which gradients exist \citep{bohte_error-backpropagation_2002, mostafa_supervised_2017, wunderlich_event-based_2021, klos_smooth_2023}.
This approach, however, requires additional mechanisms to create or remove spikes, as well as dealing with silent neurons, where the spike time is not defined.
Finally, one can replace the gradient with a suitable surrogate \citep{bohte_error-backpropagation_2011, courbariaux_binarized_2016, esser_convolutional_2016, zenke_superspike_2018, bellec_long_2018, neftci_surrogate_2019, eshraghian_training_2023}.
In this article, we focus on the latter.
\Ac{SG} approaches are empirically successful and not limited to changing the timing of existing spikes while working with non-differentiable neuron models.

However, at present \acp{SG} are a heuristic that lacks a theoretical foundation.
Consequently, we still do not understand why \ac{SG} descent works well in practice \citep{zenke_remarkable_2021}.
We do not know if it is optimal in any sense, nor do we know if there are better strategies for updating the weights.
To address this gap in our understanding we analyze the commonalities and differences of \acp{SG} with theoretically well-founded approaches based on stochastic networks.
Specifically, we focus on \acp{SPM} \citep{neftci_surrogate_2019,jang_introduction_2019} and
the more recently proposed \ac{stochAD} framework \citep{arya_automatic_2022}.
\Acp{SPM} typically rely on stochasticity to smooth out the optimization landscape in expectation to enable gradient computation on this smoothed loss.
In this article, we particularly focus on neuron models with escape noise \citep{gerstner_neuronal_2014} which are commonly used to smooth the non-differentiable spikes in expectation and for which exact gradients are computable \citep{pfister_optimal_2006,brea_matching_2013}.
However, extending \acp{SPM} to multi-layer neural networks has been difficult because they preclude using \ac{BP} and hence efficient gradient estimation requires additional approximations \citep{gardner_learning_2015}.
In contrast, \ac{stochAD} is a recently developed framework for \ac{AD} in programs with discrete randomness, i.e., with discrete random variables.
While this opens the door for \ac{BP} or other recursive gradient computation schemes, the framework has not yet been applied to \acp{SNN}.

Here, we jointly analyze the above methods, elucidate their relations, provide a rigorous theoretical foundation for \ac{SG}-descent in stochastic \acp{SNN}, and discuss their implications for deterministic \acp{SNN}.
We first provide some background information on each of the above methods before starting our analysis with the case of a single Perceptron.
From there, we move to the more complex case of \acp{MLP}, where we elaborate the theoretical connection between \acp{SG} and \ac{stochAD}, and end with multi-layer networks of \ac{LIF} neurons. Finally, we substantiate the theoretical results with empirical simulations.

 \section{Background on \aclp{SG}, \aclp{SPM}, and \acl{stochAD}}
\label{sec:analyzed_methods}
Before analyzing the relation between \acp{SG}, gradients of the log-likelihood in \acp{SPM}, and \ac{stochAD} in Section~\ref{sec:theoretical_results}, we briefly review these methods here.
While their common goal is to compute reasonable weight updates, the first two approaches are specifically designed for training \acp{SNN}. The third approach aims to compute unbiased gradients of arbitrary functions with discrete randomness based on \ac{AD}.

\paragraph{Smoothed probabilistic models.} \acp{SPM} are based on stochastic networks in which the gradients are well-defined in expectation \citep{neftci_surrogate_2019, jang_introduction_2019}.
Typically such networks consist of a noisy neuron model, such as the \ac{LIF} neuron with escape noise \citep{gerstner_neuronal_2014} and a probabilistic loss function.
On the one hand, this includes models that require optimizing the log-likelihood of the target spike train being generated by the current model \citep{pfister_optimal_2006}.
However, this method only works effectively without hidden units.
On the other hand, \acp{SPM} also comprises models that follow a variational strategy \citep{brea_matching_2013}, which makes them applicable to networks with a single hidden layer.
In general, the gradients that are computed within the \ac{SPM} framework are given by
\begin{eqnarray*}
    \frac{\partial}{\partial w}\mathbb{E}\left[\mathcal{L}(w)\right]~,
\end{eqnarray*}
where $\mathcal{L}$ is the loss and $w$ is an arbitrary model parameter.
The computational cost associated with evaluating the expected value of the loss $\mathbb{E}\left[\mathcal{L}\right]$ precludes training multi-layer networks in practice.
This is because \acp{SPM} lack support for \ac{AD}, as we will explain in detail in Section~\ref{sec:SPMs_limited_to_shallow}.

\paragraph{Surrogate gradients.}
\acp{SG} are a heuristic that relies on a continuous relaxation of the non-differentiable spiking activation function that occurs when computing gradients in \acp{SNN} \citep{neftci_surrogate_2019}, and is commonly applied in deterministic networks.
To that end, one systematically replaces the derivative of the hard threshold by a \ac{SD}, also called pseudo-derivative \citep{bellec_long_2018}, when applying the chain rule.
The result then serves as a \emph{surrogate} for the gradient.
For example, when computing  the derivative of the spike train $S$ with respect to a weight $w$, the problematic derivative of a Heaviside $\nicefrac{\partial H (u-\theta)}{\partial u}$ is replaced by a \ac{SD}
\begin{eqnarray}
    \frac{\widetilde{\partial}S}{\partial w}\leftarrow \underbrace{\frac{\partial \sigma_{\beta_\mathrm{SG}}(u-\theta)}{\partial u}}_{\mathrm{SD}}\frac{\partial u}{\partial w}~,
    \label{eq:det_SG}
\end{eqnarray}
where the tilde denotes the introduction of the surrogate. $H(\cdot)$ denotes the Heaviside function, $u$ is the membrane potential, $\theta$ is the firing threshold, and $\sigma_{\beta_\mathrm{SG}}$ is a differentiable function parametrized by $\beta_\mathrm{SG}$ used to define the \ac{SD}.
Different functions are used in practice.
For instance, the derivative of rectified linear functions or the arctangent function have been used successfully as \acp{SD} \citep{bohte_error-backpropagation_2002, esser_convolutional_2016, bellec_long_2018, hammouamri_learning_2023}, whereas SuperSpike \citep{zenke_superspike_2018} used a scaled derivative of a fast sigmoid $h(x) = \frac{1}{\left(\beta_\mathrm{SG}|x|+1\right)^2}$ with $\beta_\mathrm{SG}$ controlling its steepness.
However, \ac{SNN} training is robust to the choice of the \ac{SD} \citep{zenke_remarkable_2021, herranzcelotti_stabilizing_2024}.

\paragraph{Stochastic surrogate gradients.} In practice, \acp{SG} are most commonly used for training deterministic \acp{SNN} and have rarely been used in stochastic \acp{SNN}, with some notable exceptions \citep{gardner_learning_2015}.
Here we define stochastic \ac{SG} descent analogously to the deterministic case as a continuous relaxation of the non-differentiable spiking activation function applied to networks of stochastic neurons with escape noise \citep{gerstner_spiking_2002, gerstner_neuronal_2014}.
In these networks, spike generation is stochastic in the forward path, whereby an escape noise function yields the neuronal firing probability.
Equivalently, one can formulate stochastic neurons as neurons with a stochastic threshold $\Theta$ \citep{gerstner_spiking_2002}.
For the backward path, one considers the derivative of a differentiable function of the membrane potential.
Analogous to the deterministic case, the derivative of a spike train is hence given as
\begin{equation}
    \frac{\widetilde{\partial}S}{\partial w}\leftarrow \underbrace{\frac{\partial \sigma_{\beta_\mathrm{SG}}(u-\bar\Theta)}{\partial u}}_{\mathrm{SD}}\cdot\frac{\partial u}{\partial w}~,
    \label{eq:stochSG}
\end{equation}
where the mean threshold $\bar\Theta$ is equivalent to $\theta$ in the deterministic case (for details see Methods Section~\ref{sec:spike_generation}).
In stochastic binary neural networks, the above notion of \acp{SG} is known as the \ac{STE} \citep{hinton_lectures_2012, bengio_estimating_2013-1, courbariaux_binarized_2016, yin_understanding_2018}.

\paragraph{Stochastic automatic differentiation.}
To compute derivatives efficiently, popular \ac{AD} algorithms rely on the chain rule.
The required terms can be calculated in forward or backward mode, with the latter being called \ac{BP} \citep{marschall_unified_2019}.
To extend this to exact \ac{AD} in programs with discrete randomness, which otherwise preclude analytical gradient computation, the \ac{stochAD} framework \citep{arya_automatic_2022} introduces stochastic derivatives.
This framework provides an unbiased and composable solution to compute derivatives in discrete stochastic systems that does not suffer from the high variance found in methods that require sampling and averaging such as e.g. finite differences methods (cf. Fig.~\ref{fig:existing_frameworks} and Section~\ref{sec:SG_in_stochastic_derivatives} for a comparison).
To deal with discrete randomness, stochastic derivatives consider not only infinitesimally small continuous changes but also finite changes with infinitesimally small probability.
As defined by \citet{arya_automatic_2022}, a stochastic derivative of a random variable $X(p)$ consists of a triple $(\delta,w,Y)$.
Here $\delta$ is the “almost sure” derivative, meaning the derivative of the continuous part, $w$ is the derivative of the probability of a finite change and $Y$ is the alternate value, i.e. the value of $X(p)$ in case of a finite jump.
Although stochastic derivatives are suitable for forward mode \ac{AD}, they cannot be used directly for \ac{BP} as of now.
This is because the application of the chain rule would require the derivatives to be scalars, whereas they consist of a triple.
The \ac{stochAD} framework outlines a way to convert them to a single scalar value, which the authors call the smoothed stochastic derivative.
This transformation renders the framework compatible with \ac{BP} albeit at the cost of introducing bias.
Given a stochastic derivative, its smoothed stochastic derivative is
\begin{eqnarray}
    \label{eq:smoothed_ssd}
    \tilde{\delta} = \mathbb{E}[\delta + w(Y-X(p))|X(p)]
\end{eqnarray}
for one realization of the random variable $X(p)$.
\citet{arya_automatic_2022} also showed that the smoothed stochastic derivatives recover the \ac{STE} \citep{hinton_lectures_2012, bengio_estimating_2013-1} for a Bernoulli random variable in their example A.8.

\section{Analysis of the relation between \acp{SG}, \acp{SPM}, and \ac{stochAD} for direct training of \acp{SNN}}
\label{sec:theoretical_results}
To fathom the theoretical foundation of \ac{SG} learning, we focused on the theoretically well-grounded \acp{SPM} and the recently proposed \ac{stochAD} framework.
Because both approaches assume stochasticity whereas \acp{SG} are typically applied in deterministic settings, we focus our analysis on stochastic networks (cf. Section~\ref{sec:analyzed_methods}).
We will later discuss deterministic networks as a special case.

To keep our analysis general and independent of the choice of the loss function $\mathcal{L}$, we consider the Jacobian $ \nabla_w y $ defined at the network's output $y$ where $w$ are the trainable parameters or weights.
For simplicity and without loss of generality, we consider networks with only one output such that the above is equivalent to studying
\begin{equation*}
    \nabla y = \left(\frac{\partial}{\partial w_1}y, \dots, \frac{\partial }{\partial w_n}y\right) ~ .
\end{equation*}

To further ease the analysis, we start with binary Perceptrons, thereby neglecting all temporal dynamics and the reset of conventional spiking neuron models, while retaining the essential binary spike generation process (see Fig.~\ref{fig:SG_are_not_equivalent}A, Methods Section~\ref{sec:spike_generation}).
We begin our comparison by examining a single neuron before moving on to \acp{MLP}.
We defer the discussion of \ac{LIF} neurons to Section~\ref{sec: lif_neurons}.

\subsection{Analysis of the binary Perceptron}
\label{sec:single_neurons}

Before we compare the different methods of computing gradients or \acp{SG}, we define our deterministic and stochastic Perceptron here.

\paragraph{Deterministic Perceptron.} The deterministic Perceptron is defined as
\begin{eqnarray}
    u &=& W^T x + b
    \label{eq:det_Perceptron_mem_pot} \\
    y &=& H(u-\theta)~,
    \label{eq:det_Perceptron_spike}
\end{eqnarray}
where $H(\cdot)$ is again the Heaviside step function and $u$ is the membrane potential, which depends on the weights $W$, the bias $b$, the input $x$, and the firing threshold $\theta$.
The Jacobian is related to the gradient via
\begin{equation*}
    \frac{\partial \mathcal{L}}{\partial w_i} = \frac{\partial \mathcal{L}}{\partial y} \frac{\partial y}{\partial w_i}~,
\end{equation*}
where the problematic derivative of the non-differentiable Heaviside function appears in $\frac{\partial y}{\partial w_i}$ with $w_i$ being the $i$th weight.
When computing \acp{SG} the derivative of the Heaviside function is replaced with the corresponding \ac{SD}
\begin{equation}
    \frac{\widetilde{\partial} }{\partial w_i} y(u-\theta) \leftarrow \frac{\partial}{\partial w_i}\sigma_{\beta_\mathrm{SG}}(u-\theta) ~ .
    \label{eq:SD_Perceptron}
\end{equation}
Here we chose the \ac{SD} as the derivative of a sigmoid with steepness $\beta_\mathrm{SG}$, hence $\sigma_{\beta_\mathrm{SG}}(u-\theta) = \frac{1}{1+\exp(-\beta_\mathrm{SG} (u-\theta))}$.

\begin{figure}[tbp]
    \centering
    \includegraphics[]{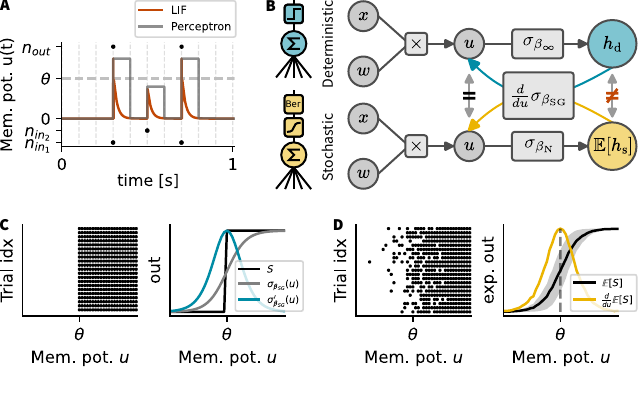}
    \caption{\textbf{\Acp{SD} are equivalent to derivatives of expected outputs in \acp{SPM} and smoothed stochastic derivatives in binary Perceptrons.}
        \textbf{(A)}~Membrane potential dynamics of an \ac{LIF} neuron (maroon) in comparison to the Perceptron.
        When input spikes (from input neurons $n_{\mathrm{in}_1}$ and $n_{\mathrm{in}_2}$) are received, they excite the \ac{LIF} neuron which causes the membrane potential to increase.
        Once it reaches the threshold, output spikes $n_\mathrm{out}$ are emitted.
        In the limit of a large simulation time step ($dt \gg \tau_{mem}$) and appropriate scaling of the input currents, the \ac{LIF} neuron approximates a Perceptron receiving time-locked input (gray line).
        \textbf{(B)}~Left: The simplified computational graph of a deterministic (blue) and a stochastic (yellow) Perceptron.
        Right: Forward pass in a deterministic (top) or stochastic (bottom) Perceptron.
        The colored arrows indicate that both use the derivative of $\sigma_{\beta_\mathrm{SG}}(\cdot)$ on the backward pass, which is the derivative of the expected output of the stochastic neuron in case $\beta_\mathrm{SG} = \beta_\mathrm{N}$.
        In the case of the deterministic neuron, this constitutes the \ac{SG} used instead of the non-existing derivative of the step function.
        \textbf{(C)}~Left: Network output over multiple trials in the deterministic Perceptron. Right: The \ac{SD} (blue) is the derivative of a sigmoid (gray), which is used to approximate the non-existing derivative of the step function (black).
        \textbf{(D)}~Same as (C) but for the stochastic Perceptron. Left: Escape noise leads to variability in the spike trains over trials. Right: The expected output follows a sigmoid, and we can compute the derivative (yellow) of the expected output.
    }
    \label{fig:SG_are_not_equivalent}
\end{figure}

\paragraph{Stochastic Perceptron.} To see how the above expression compares to \acp{SD} and derivatives of the expected output in \acp{SPM} in the corresponding stochastic setting, we consider the stochastic Perceptron
\begin{eqnarray}
    \nonumber
    u &=& W^T x + b \\
    p &=& f(u-\theta) = \sigma_{\beta_\mathrm{N}}(u-\theta) \label{eq:stoch_Perceptron} \\
    y &\sim& \mathrm{Ber}(p) ~. \nonumber
\end{eqnarray}
Note that the membrane potential $u$ is equivalent to the deterministic case (cf. Eq.~\eqref{eq:det_Perceptron_mem_pot}), as the models share the same input.
To model stochasticity, we add escape noise.
This means the Perceptron fires with a probability $p$, which is a function $f$ of the membrane potential.
To that end, we choose $f(\cdot)$ as the sigmoid function $\sigma_{\beta_\mathrm{N}}(u-\theta) = \frac{1}{1+\exp(-\beta_\mathrm{N} (u-\theta))}$, where $\beta_\mathrm{N}$ controls the steepness of the escape noise function \citep{brea_matching_2013}.
Importantly, there is some degree of freedom as to which escape noise function we choose.
Other common choices in the realm of spiking neurons are the exponential or error function \citep{gerstner_neuronal_2014}.
For our current choice, the sigmoid, we find that it approaches the step function in the limit $\beta_\mathrm{N} \rightarrow \infty$, and the stochastic Perceptron becomes deterministic.

\medskip
Let us now compare \ac{SG} descent in a deterministic and a stochastic unit.
To do so, we first incorporate the spike generation of the stochastic binary Perceptron into the threshold and reformulate the stochastic firing using $H(u-\Theta)$.
The stochastic threshold $\Theta$ is given by $\Theta=\theta+\xi$, where $\xi$ is randomly drawn noise from a zero-mean distribution.
This formulation is equivalent to stochastic firing as in Eq.~\eqref{eq:stoch_Perceptron} for a suitable choice of the distribution underlying $\xi$ (cf. Eq.~\eqref{eq:stoch_spikes_v2}).
The stochastic \ac{SD} is then as in Eq.~\eqref{eq:stochSG}.
Since the input and the weights are the same in both cases, both units have the same membrane potential.
Because the \ac{SD} depends only on the membrane potential the resulting weight updates are equivalent provided that both use the same $\beta_\mathrm{SG}$ (cf.\ Eqs.~\eqref{eq:det_SG} and \eqref{eq:stochSG}).
Thus, the resulting \acp{SD} in the stochastic and deterministic Perceptron are equivalent although the outputs of the two units are generally different (Fig.~\ref{fig:SG_are_not_equivalent}C, D), and hence the overall weight update is different.

We now turn to comparing \acp{SG} to \acp{SPM}.
In \acp{SPM}, the idea is to compute the derivative of the expected loss at the output to smooth out the discrete spiking non-linearity \citep{pfister_optimal_2006}.
In our setup, this amounts to computing the expected output of the stochastic Perceptron,
\begin{equation*}
    \mathbb{E}[y] = \sigma_{\beta_\mathrm{N}}(u-\theta),
\end{equation*}
and taking its derivative
\begin{equation}
    \frac{\partial}{\partial w_i}\mathbb{E}[y] = \frac{\partial}{\partial w_i}\sigma_{\beta_\mathrm{N}}(u-\theta)~.
    \label{eq:bin_stochgrad}
\end{equation}

We note that the right-hand sides of Expressions~\eqref{eq:bin_stochgrad} and~\eqref{eq:SD_Perceptron} are the same when setting $\beta_\mathrm{N} = \beta_\mathrm{SG}$.
Thus, the derivative of the expected output of the stochastic Perceptron is equivalent to the \ac{SD} of the output of the Perceptron, if the \ac{SD} is chosen such that it matches the derivative of the neuronal escape noise (see Fig.~\ref{fig:SG_are_not_equivalent}B).

Finally, we compare the above findings to the derivative obtained from the \ac{stochAD} framework.
To that end, we first apply the chain rule given one realization of $y$
\begin{eqnarray}
    \frac{\partial y}{\partial w_i} &=& \frac{\partial } {\partial p} \mathrm{Ber}(p)\frac{\partial}{\partial w_i}p
    \label{eq:stochAD_example}
\end{eqnarray}
and use the smoothed stochastic derivative of a Bernoulli random variable following the steps of \citet{arya_automatic_2022}.
According to their example A.8 \citep{arya_automatic_2022}, the right stochastic derivative for a Bernoulli random variable is given by $(\delta_R, w_R, Y_R) = (0, \frac{1}{1-p}, 1)$ if the outcome of the random variable was zero ($\mathrm{Ber}(p)=0$) and zero otherwise.
The left stochastic derivative is given by $(\delta_L, w_L, Y_L) = (0, \frac{-1}{p}, 0)$ if $\mathrm{Ber}(p)=1$ and also zero otherwise.
The corresponding smoothed versions are $\widetilde{\delta}_R = \frac{1}{1-p} \cdot \mathbf{1}_{X(p)=0}$ and $\widetilde{\delta}_L = \frac{1}{p} \cdot \mathbf{1}_{X(p)=1}$ (cf. \citet{arya_automatic_2022} or Eq.~\eqref{eq:smoothed_ssd}) .
Since every affine combination of the left and right derivatives is a valid derivative, we can use $(1-p)\widetilde{\delta}_R + p\widetilde{\delta}_L = 1$ as the smoothed stochastic derivative of the Bernoulli random variable.
Hence, as previously seen in \citet{arya_automatic_2022}, the smoothed stochastic derivative of a Bernoulli can be equal to 1.
This results in
\begin{eqnarray*}
    \frac{\partial y}{\partial w_i} &\leftarrow& 1 \cdot \frac{\partial}{\partial w_i} \sigma_{\beta_\mathrm{N}}(u-\theta)
\end{eqnarray*}
when inserted into Eq.~\eqref{eq:stochAD_example}.
Yet again, we obtain the same right-hand side as above.
It is worth noting that in the comparison with \ac{SG} descent, we consider not only the smoothed stochastic derivative of the Bernoulli random variable but also its composition with the derivative of the firing probability, or equivalently, the neuron's escape noise function.
This composition is necessary because we require a surrogate for the ill-defined derivative of the spike train with respect to the membrane potential $\frac{\partial S}{\partial u}$ that is also valid in the deterministic case.

In summary, when analyzing different ways of computing or approximating derivatives, the resulting expressions are identical in single units, if the \ac{SD} matches the derivative of the escape noise function of the stochastic neuron model.
In the single-neuron case, this match is equivalent to probabilistic smoothing of the loss, a well-known fact in the literature \citep{pfister_optimal_2006}, which allows computing exact gradients of the expected value.
Nevertheless, the resulting weight updates are generally different in the deterministic and stochastic cases because these expressions also depend on the neuronal output, which is different in the deterministic and stochastic neurons.
We will see in the next section, which equivalences are broken in multi-layer networks due to these and other notable differences.

\subsection{Analysis of the multi-layer Perceptron}
\label{sec:deep_nets}
To analyze the relation of the different methods in the multi-layer setting, we begin by examining \acp{SPM}, which lack support for \ac{AD} and thus an efficient algorithm to compute gradients.
We then further discuss how \ac{stochAD} provides smooth stochastic derivatives equivalent to \acp{SG} in multi-layer networks.

\subsubsection{Output smoothing in multi-layer \acp{SPM} precludes efficient gradient computation}
\label{sec:SPMs_limited_to_shallow}
\acp{SPM} lack support for \ac{AD}, because they smooth the expected loss landscape through stochasticity and therefore require the calculation of expected values at the network output.
While output smoothing allows the implementation of the finite difference algorithm, this algorithm does not scale to large models and is therefore of little practical use for training \acp{ANN} \citep{werfel_learning_2004,lillicrap_backpropagation_2020}.
The application of \ac{AD}, however, requires differentiable models, like standard \acp{ANN}, so that the chain rule can be used to decompose the gradient computation into simple primitives.
This composability is the basis for efficient recursive algorithms like \ac{BP} and \ac{RTRL} \citep{marschall_unified_2019}.

\begin{figure}[tbp]
    \centering
    \includegraphics{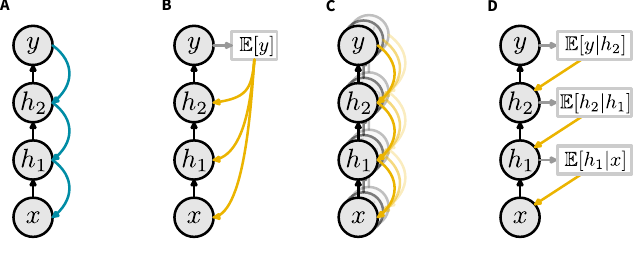}
    \caption{
        \textbf{Derivative computation in \acp{MLP}.} Schematic of an example network for which (surrogate) derivatives are computed according to different methods.
        The colored arrows indicate where partial derivatives are calculated.
        \textbf{(A):}~\Ac{SG} descent relies on the chain rule for efficient gradient computation in a deterministic \ac{MLP}.
        Thus, the derivative of the output with respect to a given weight is factorized into its primitives, which are indicated by the colored arrows.
        \textbf{(B)}~\acp{SPM} approach the problem of non-differentiable spike trains by adding noise and then smoothing the output based on its expected value.
        Since this method does not allow the use of the chain rule, the derivative for each weight must be computed directly.
        \textbf{(C)}~The derivative and the expected value are not interchangeable, which makes this option mathematically invalid.
        Furthermore, it is not possible to achieve the necessary smoothing using the expected value after such an interchange.
        \textbf{(D)}~Smoothed stochastic derivatives in \ac{stochAD} use the expected value of each node to compute the derivative.
        However, the method relies on expectation values conditioned on the activity of a specific forward pass.
    }
    \label{fig:maths}
\end{figure}

To see why \acp{SPM} do not support \ac{AD}, let us consider a simple example network:
Let $y$ be the output of a binary neural network with input $x$ and two hidden layers with activities $h_1$, $h_2$ (Fig.~\ref{fig:maths}).
The output $y$ has the firing probability $p_y = \sigma_{\beta_\mathrm{N}}(w_y^Th_2)$ as in Eq.~\eqref{eq:stoch_Perceptron}, and the hidden layers have the firing probabilities $p_1=\sigma_{\beta_\mathrm{N}}(w_1^Tx)$ and $p_2 = \sigma_{\beta_\mathrm{N}}(w_2^Th_1)$ respectively.
We are looking for a closed-form expression of the derivative of the expected output for each parameter, e.g., $\frac{\partial }{\partial w_1}\mathbb{E}[y]$ for weight $w_1$ (Fig.~\ref{fig:maths}B).
In a deterministic and differentiable network, one can use the chain rule to split the expression into a product of partial derivatives as $\frac{\partial }{\partial w_1}y = \frac{\partial y}{\partial p_y}\frac{\partial p_y}{\partial h_2}\frac{\partial h_2}{\partial p_2}\dots\frac{\partial p_1}{\partial w_1}$ (see Fig. ~\ref{fig:maths}A).
However, this is not possible for \acp{SPM} because $\frac{\partial }{\partial w_1}\mathbb{E}[y] \ne \mathbb{E}\left[\frac{\partial }{\partial w_1}y\right]$, where the right-hand side would involve the derivative of the non-differentiable binary output.

Even if we could exchange the expectation value and the derivative, we would still be faced with the fact that the expectation of a product is usually not equal to the product of expectation values, unless the factors are independent (Fig.~\ref{fig:maths}C), hence
\begin{eqnarray*}
    \mathbb{E}\left[\frac{\partial }{\partial p_y}y\frac{\partial }{\partial h_2}p_y\frac{\partial }{\partial p_2}h_2\dots\frac{\partial}{\partial w_1} p_1\right]
    &\ne& \mathbb{E}\left[\frac{\partial }{\partial p_y}y\right]\mathbb{E}\left[\frac{\partial }{\partial h_2}p_y\right]\mathbb{E}\left[\frac{\partial }{\partial p_2}h_2\right]\dots\mathbb{E}\left[\frac{\partial }{\partial w_1}p_1\right]\\
    &\ne& \frac{\partial }{\partial p_y}\mathbb{E}\left[y\right]\frac{\partial }{\partial h_2}\mathbb{E}\left[p_y\right]\frac{\partial }{\partial p_2}\mathbb{E}\left[h_2\right]\dots\frac{\partial }{\partial w_1}\mathbb{E}\left[p_1\right]~.
\end{eqnarray*}
Clearly, in neural networks, the factors are not independent, because the activity of downstream neurons depends on the activity of their upstream partners.
Thus, it is not obvious how to compute gradients in multi-layer \ac{SPM} networks.
We will see that \ac{stochAD} suggests sensible solutions to the aforementioned problems which ultimately justify why we \emph{can} in fact do some of the above operations.
Consequently, in the following, we will only consider \acp{SG} and \ac{stochAD}, which support \ac{BP}.

\subsubsection{\Acl{stochAD} bestows theoretical support to \aclp{SG}}
\label{sec:SG_in_stochastic_derivatives}
Here we show how smoothed stochastic derivatives for stochastic binary \acp{MLP} relate to \acp{SG}.
\citet{arya_automatic_2022} defined the smoothed stochastic derivative (repeated in Eq.~\eqref{eq:smoothed_ssd} for convenience) for one realization of a random variable $X(p)$, where the discrete random variables are usually Bernoulli random variables in stochastic binary \acp{MLP}.

\begin{figure}[tbp]
    \centering
    \includegraphics[]{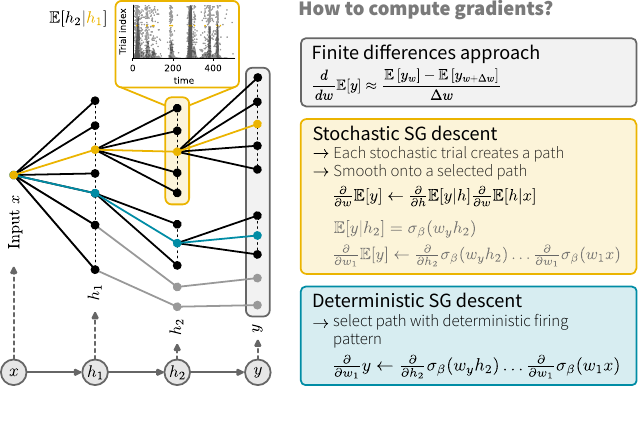}
    \caption{
    \textbf{\acp{SG} correspond to smoothed stochastic derivatives in stochastic \acp{SNN}.}
    The tree illustrates the discrete decisions associated with the binary spike generation process at different points over different layers of a stochastic \ac{MLP}.
    A single forward pass in the network corresponds to a specific path through the tree which yields a specific set of spike trains.
    Another forward pass will result in a different path and spike patterns.
    Computing gradients using \textbf{finite differences} requires randomly sampling paths from the network and evaluating their averaged loss before and after a given weight perturbation.
    Although this approach is unbiased for small perturbations, the random path selection results in high variance.
    Furthermore, that approach is not scalable to large networks.
    \textbf{Stochastic \ac{SG} descent} is equivalent to smoothed stochastic derivatives in the \ac{stochAD} framework.
    To compute the gradient, we roll out the network once and sample a random path in the tree which we now keep fixed (yellow).
    At each node, we then compute the expected output given the fixed activation of the previous layer $\mathbb{E}[h_i|h_{i-1}]$, which yields a low-variance estimate (see inset: spike raster, selected trial shown in yellow, spike trains of other trials in gray, expectation shown as shaded overlay).
    By choosing a surrogate function that matches the escape noise process, both methods give the same derivative for a spike with respect to the membrane potential.
    \textbf{Deterministic \ac{SG} descent} can be seen as a special case in which the random sampling of the path is replaced by a point estimate given by the deterministic roll-out (blue).
    }
    \label{fig:existing_frameworks}
\end{figure}

To gain an intuitive understanding of the essence of smoothed stochastic derivatives, we consider a single realization of a stochastic program, which, in our case, is a stochastic \ac{MLP}.
In other words, we run a single stochastic forward pass in the \ac{MLP} and condition on its activity.
At each point in time, each neuron either emits a one or a zero, i.e., a spike or none.
Thus, we can think of all these binary decisions as edges in a tree, with each node representing the spike pattern of all units in a given layer.
Any roll-out of a forward pass then corresponds to a particular randomly chosen path through this tree (Fig.~\ref{fig:existing_frameworks}).
These paths originate from the same root for a given input $x$, yet the stochasticity in all of them is independent.
In this tree of possible spike patterns, it is easy to understand how different methods for gradient estimation work.

Let us first consider the finite differences method.
In this approach, the loss is evaluated once using the parameters $w$ and once using $w+\Delta w$, either considering a single trial or averaging over several independent trials.
Hence, one randomly samples paths in the tree of spike patterns and compares their averaged output before and after weight perturbation.
Since the randomness of the sampled paths is ``uncoupled,'' the finite difference approach results in high variance (cf.\ Fig.~\ref{fig:existing_frameworks}, gray), which scales with the inverse of the squared perturbation $\Delta w$. However, smaller perturbations allow a more accurate gradient estimation \citep{glasserman_estimating_2003,fu_chapter_2006}.
Reducing this variance requires instead the introduction of a coupling between the sampled paths, which is at the heart of \ac{stochAD} \citep{arya_automatic_2022}.
The key idea is to condition on the previous layer's output $h_{i-1}$ and then consider all possible spike patterns for the current layer $\mathbb{E}[h_i|h_{i-1}]$.
Thus, we can consider the expected layer activity, given a randomly chosen path, e.g., the yellow path in Fig.~\ref{fig:existing_frameworks}.
For this path, the derivatives now need to be smoothed at each node.
To do so, we compute the expectation in each layer conditioned on the activity of the previous layer along the selected path.
After smoothing, it is possible to compute the path-wise derivative along a selected path with activity $h_1^*, h_2^*, y^*$:
\begin{equation*}
    \frac{\partial}{\partial w_1} \mathbb{E}[y] = \frac{1}{n_\mathrm{paths}}\sum_{\mathrm{paths}} \frac{\partial }{\partial p_y}\mathbb{E}[y|h_2^*]\frac{\partial }{\partial h_2}\mathbb{E}[p_y|h_2^*]\frac{\partial }{\partial p_2}\mathbb{E}[h_2|h_1^*]\cdots\frac{\partial }{\partial w_1}\mathbb{E}[p_1|x]~.
\end{equation*}
Here we averaged over all possible paths, i.e., all possible combinations of the activities $h_1^*, h_2^*$.
In practice, it is rarely possible to average over all possible combinations and one instead uses a Monte Carlo estimate, which still yields significantly lower variance than other schemes and can be computed efficiently using \ac{BP} for a single path per update.

Given the above method for computing smoothed stochastic derivatives, we are now in the position to understand their relationship to \acp{SG}.
Since we condition on a specific path in the tree of all possible spike patterns, we only compute derivatives of the expected output $h_i$ conditional on the output of the previous layer $h_{i-1}$ according to the chosen path at each node.
Such an approach exactly corresponds to treating each node as a single unit like in the previous Section~\ref{sec:single_neurons}.
As we saw above, the derivative of the expected output of a single unit with escape noise is equivalent to the corresponding smoothed stochastic derivative and the \ac{SD} in that unit both with or without escape noise.
Furthermore, when using \acp{SG}, there is no difference in how the method is applied in single versus multi-layer cases and there is always a well-defined path to condition on.
So, \acp{SG} can also be understood as treating all units at each layer as single units.
The same is true for smoothed stochastic derivatives in the \ac{stochAD} framework.
Thus, the \ac{stochAD} framework provides theoretical support to \acp{SG} in stochastic networks and also stipulates that the \ac{SD} should be matched to mirror the derivative of the neuronal escape noise, thereby extending this matching principle from the single neuron case \citep{pfister_optimal_2006, brea_matching_2013} to the multi-layer network setting and removing some of the ambiguity associated with choosing this function in practice \citep{zenke_remarkable_2021}.

\subsubsection{Surrogate gradients in deterministic networks}
\acp{SG} are commonly used to train deterministic \acp{SNN}.
Deterministic networks show two essential differences to their stochastic counterparts which result in different weight updates from what we have learned above.
First,
while a stochastic network typically selects a different path in the tree of spike patterns in each trial, there is only one path per input and parameter set in the deterministic case.
This means that the expectation value over sampling random paths is replaced by a point estimate provided by the deterministic roll-out.
This difference introduces a selection bias in the estimation of the stochastic implementation.
However, in practice, this bias is often small (Supplementary Fig.~\ref{sfig:bias}).
Second, when training deterministic networks with \acp{SG} the slope of the \ac{SD} at threshold ($\beta_\mathrm{SG}$) is typically chosen ad-hoc.
Typical values are on the order one compared to the neuronal rheobase.
This is stark in contrast to the ``correct'' slope of the corresponding asymptotic escape noise function from taking the limit $\beta_\mathrm{N}\to\infty$.
Approaching this limit empirically with  $\beta_\mathrm{SG}$ leads to unstable training and poor performance \citep{zenke_remarkable_2021}.
The choice $\beta_\mathrm{SG} \ne \beta_\mathrm{N}$ inevitably results in different weight updates.
While this is a desired effect that allows training in the first place by avoiding the problem of zero and infinite weight updates, it is less clear whether this approximation has any undesirable off-target consequences.
To address this question we will examine the properties of \acp{SG} in deterministic networks in the next section.

\section{Analysis of \acl{SG} properties in deterministic networks}
\label{sec:sg_biased}
By design \acp{SG} deviate from the actual gradient, as they provide a non-zero surrogate in cases where the actual gradient is zero.
It is not clear a-priori what consequences such a deviation has and whether \acp{SG} are gradients at all, which can be obtained by differentiating a surrogate loss.
However, it is difficult to get a quantitative understanding when comparing either to the zero vector or to the ill-defined derivative at the point of the spike.
To take a closer look at how \acp{SG} deviate from actual gradients and the resulting consequences we now move to differentiable network models that have a well-defined gradient.
While \ac{SG} training is not required in such networks, it allows us to develop a quantitative understanding of the commonalities and differences we should expect.
This comparison also allows us to check whether \acp{SG} satisfy the formal criteria of gradients.

\subsection{Deviations of \acp{SG} from actual gradients in differentiable \acp{MLP} and the consequences}
\label{sec: bias_in_sigmoid_networks}

\begin{figure}[tbp]
    \centering
    \includegraphics[]{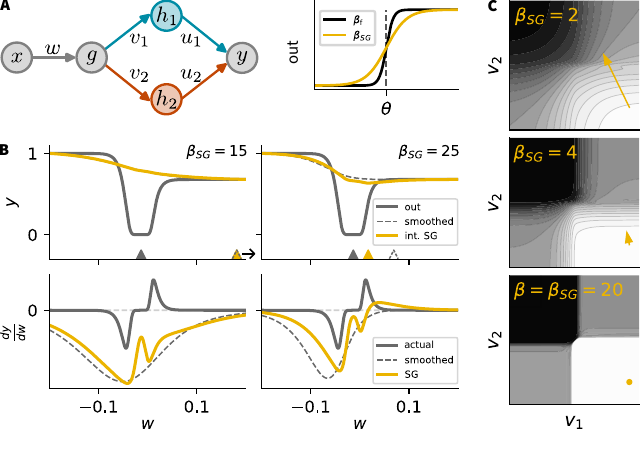}
    \caption{
        \textbf{\acp{SG} deviate from actual gradients in differentiable \acp{MLP}.}
        \textbf{(A)}~Schematic of a differentiable network (left) with sigmoid activations (right) for which we compute an \ac{SD} using the derivative of a flatter sigmoid (yellow) in contrast to the actual activation (black).
        \textbf{(B)}~Top row: Network output (solid gray), smoothed network output (dashed), and integrated \ac{SD} (yellow) as a function of $w$.
        The triangles on the x-axis indicate the minimum of the corresponding curves.
        Bottom row: Derivatives of the top row.
        Left and right correspond to a flatter ($\beta_{\mathrm{SG}}=15$) and a steeper ($\beta_\mathrm{SG}=25$) \ac{SD}, see Table~\ref{tab:sign_flip_params} for network parameters.
        Note that the actual derivative and the surrogate can have opposite signs.
        \textbf{(C)}~Heatmap of the optimization landscape along $v_1$ and $v_2$ for different $\beta_\mathrm{SG}$ values (top to bottom).
        While the actual gradient can be asymptotically zero (see yellow dot, bottom), the \ac{SD} provides a descent direction (yellow arrow), thereby enabling learning (top and middle).
    }
    \label{fig:bias_in_sigmoid_networks}
\end{figure}

We first sought to understand whether \acp{SG} point in a ``similar direction'' as the actual gradient.
Specifically, we asked whether \acp{SG} ensure sign concordance, i.e., whether they preserve the sign of the gradient components.
To investigate this question, we consider a small network with input $x$ and output $y$ (Fig.~\ref{fig:bias_in_sigmoid_networks}A) defined by
\begin{eqnarray}
    g &=& \sigma_{\beta_f}(wx) \nonumber \\
    h_1 &=& \sigma_{\beta_f}(v_1g)\nonumber \\
    h_2 &=& \sigma_{\beta_f}(v_2g) \nonumber \\
    y &=& \sigma_{\beta_f}(u_1 h_1 + u_2 h_2) ~ .
    \label{eq:example_net}
\end{eqnarray}
The network parameters are denoted by $w$, $v_1$, $v_2$, $u_1$, and $u_2$, while $g$, $h_1$, and $h_2$ are the hidden layer activations.
This network has a well-defined gradient and provides a minimal working example.
To study the effect of computing a \ac{SD}, we replace the sigmoid parameterized with $\beta_f$ used in the forward pass by a surrogate function with $\beta_{\mathrm{SG}}$ which is used to compute the \ac{SD} in the backward pass.
Hence the \ac{SD} for a steeper sigmoid is given by
\begin{equation}
    \label{eq:sigmoid_SG}
    \frac{\widetilde{\partial}}{\partial w}\sigma(\beta_f \cdot wx) \leftarrow x\cdot\beta_{\mathrm{SG}}\cdot \sigma(\beta_{\mathrm{SG}} \cdot wx)\cdot (1-\sigma(\beta_{\mathrm{SG}} \cdot wx))
\end{equation}
with $\beta_f > \beta_{\mathrm{SG}}$.
As before, the deterministic binary Perceptron corresponds to the limiting case $\beta_f\to\infty$.

\begin{table}[tbp]
    \centering
    \caption{\textbf{Parameter values for sign flip example.}
        Parameter values for the network in Fig.~\ref{fig:bias_in_sigmoid_networks}A, with input $x=1$, which serve as an example, that \acp{SG} can have the opposite sign of the actual gradient and thus point towards the opposite direction.
        Therefore, we cannot guarantee the \ac{SG} to align with the actual gradient.
    }
    \vspace{8pt}
    \begin{tabular}{l|ccccccc|c}
        \toprule
        \textbf{Parameter} & $w$ & $v_1$ & $v_2$ & $u_1$ & $u_2$ & $\beta_\mathrm{f}$ & $\beta_{\mathrm{SG}}$ & Input $x$ \\
        \textbf{Value}     & 0   & 0.05  & 0.1   & 1     & -1    & 100                & 25                    & 1         \\
        \bottomrule
    \end{tabular}
    \label{tab:sign_flip_params}
\end{table}

To investigate the differences between the \ac{SG} and the actual gradient, we are particularly interested in the derivative of the output with respect to the hidden layer weight $w$ (cf. Fig.~\ref{fig:bias_in_sigmoid_networks}A).
The partial derivative of the output with respect to $w$ is given as
\begin{eqnarray*}
    \frac{\partial }{\partial w}y &=& \underbrace{\frac{\partial y}{\partial h_1}\frac{\partial h_1}{\partial g}\frac{\partial g}{\partial w}}_\mathrm{blue path} + \underbrace{\frac{\partial y}{\partial h_2}\frac{\partial h_2}{\partial g}\frac{\partial g}{\partial w}}_\text{maroon path}~.
\end{eqnarray*}
Now inserting the \acp{SD} using Eq.~\eqref{eq:sigmoid_SG} leads to
\begin{eqnarray}
    \nonumber
    \frac{\widetilde{\partial} }{\partial w}y &=&
    \left(\frac{\widetilde{\partial }y}{\partial h_1}\frac{\widetilde{\partial }h_1}{\partial g}
    + \frac{\widetilde{\partial }y}{\partial h_2}\frac{\widetilde{\partial }h_2}{\partial g}\right)\frac{\widetilde{\partial }g}{\partial w}\\
    &\leftarrow& \underbrace{\big( u_1 v_1 \cdot \sigma_{\beta_{\mathrm{SG}}}'(v_1g)+ u_2 v_2 \cdot \sigma_{\beta_{\mathrm{SG}}}'(v_2g)\big)}_{\text{responsible for sign flip}} \cdot \underbrace{\beta_{\mathrm{SG}}^3 \cdot\sigma_{\beta_{\mathrm{SG}}}'(u_1h_1 + u_2h_2) \cdot \sigma_{\beta_{\mathrm{SG}}}'(wx)}_{\text{positive factor}}\cdot x~,
    \label{eq:sign_flip}
\end{eqnarray}
where only the first factor in Eq.~\eqref{eq:sign_flip}, which consists of two terms, determines the relative sign of the \ac{SD} with respect to the actual derivative.
This is because the second factor, as well as $\beta_{\mathrm{SG}}$, are always positive.
Furthermore, the derivative of the sigmoid is always positive independently of the choice of $\beta_\mathrm{f}$ or $\beta_{\mathrm{SG}}$.
Finally, $x$, the input data, does not change its sign dependent on $\beta_{\mathrm{SG}}$.
However, the first factor can change its sign, since it is a summation of two nonlinear functions with changed hyperparameters $\beta_\mathrm{f}$ or $\beta_{\mathrm{SG}}$ and different weights, which may be negative.
For instance, when we use specific parameter values (given in Table~\ref{tab:sign_flip_params}) in Eq.~\eqref{eq:sign_flip}, the \ac{SD} has the opposite sign of the actual derivative (Fig.~\ref{fig:bias_in_sigmoid_networks}B).
Thus, already in this simple example, there is no guarantee that the sign of the \ac{SG} is preserved with respect to the actual gradient.
As a consequence following the \ac{SG} will not necessarily find parameter combinations that correspond to a minimum of the loss.

\subsection{Surrogate gradients are not gradients of a surrogate loss}
\label{sec: SG_are_not_gradients}

\begin{figure}[tpb]
    \centering
    \includegraphics[]{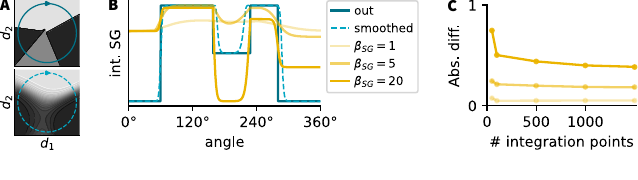}
    \caption{
        \textbf{\acp{SG} are not gradients.}
        \textbf{(A)}~Heat map of the network output while moving along two random directions in the parameter space of the example network with step activation (top) and sigmoid activation (bottom) (see Fig.~\ref{fig:bias_in_sigmoid_networks} A).
        The circles indicate a closed integration path through parameter space, starting at the arrowhead.
        \textbf{(B)}~Integral values of \acp{SD} as a function of the angle along the closed circular path shown in (A).
        Different shades of yellow correspond to different values of $\beta_\mathrm{SG}$.
        The blue lines correspond to the actual output of the network with step activation function (solid) or sigmoid activation (dashed).
        The integrated actual derivative of the network with sigmoid activation matches the output (dashed line) and is thus not visible in the plot.
        \textbf{(C)}~Absolute difference between actual loss value and integrated \ac{SD} as function of the number of integration steps.
        The numerical integrals converge to finite values.
        Thus the observed difference is not an artifact of the numerical integration.
    }
    \label{fig:SG_are_not_grad_of_loss}
\end{figure}

Given the above insight, we wondered whether a \ac{SG} can be understood as the gradient of a surrogate loss that is not explicitly defined.
To answer this question, we note that if, and only if, the \ac{SG} is the gradient of a scalar function, i.e., corresponds to a conservative field, then integrating over any closed path must yield a zero integral.
To check this, we considered the approximate Jacobian obtained using the \acp{SD} in the above example and numerically computed the integral over a closed circular path parameterized by the angle $\alpha$ in a two-dimensional plane in parameter space for different values of $\beta_\mathrm{SG}$
\begin{eqnarray*}
    I_\mathrm{SG} &=& \int_{0^\circ}^{360^\circ} \frac{\widetilde d y(\theta_\alpha)}{d \theta_\alpha}\frac{d \theta_\alpha}{d\alpha} d\alpha ~,
\end{eqnarray*}
where the network parameters at any position are given by $\theta_\alpha$ (Fig.~\ref{fig:SG_are_not_grad_of_loss}A, B; Methods Section~\ref{sec:met:example_net}).
We found that integrating the \ac{SD} did not yield a zero integral, whereas using the actual derivatives resulted in a zero integral as expected.
Importantly, this difference was not explained by numerical integration errors due to the finite step size (Fig.~\ref{fig:SG_are_not_grad_of_loss}C).
Thus \acp{SG} cannot be understood as gradients of a surrogate loss.

\section{From Perceptrons to \ac{LIF} neurons}
\label{sec: lif_neurons}
In our above treatment, we focused on binary Perceptrons for ease of analysis.
In the following, we show that our findings readily generalize to networks of \ac{LIF} neurons.
To that end, we consider \ac{LIF} neurons in discrete time which share many commonalities with the binary Perceptron (see also Fig.~\ref{fig:SG_are_not_equivalent}A).
To illustrate these similarities let us consider a single \ac{LIF} neuron with index $i$ described by the following discrete-time dynamics:
\begin{eqnarray}
    I_i[n+1] &=& \lambda_\mathrm{s} I_i[n] + \sum_j w_{ij} S_j[n]
    \label{eq:current}\\
    U_i[n+1] &=& \left(\lambda_\mathrm{m}U_i[n]+\left(1-\lambda_\mathrm{m}\right)I_i[n]\right)\left(1-S_i[n]\right)
    \label{eq:potential}\\
    S_i[n] &=& H(U_i[n]-\theta)~,
    \label{eq:spike_gen}
\end{eqnarray}
where $w_{ij}$ describes the synaptic weights between the neuron $i$ and the input neuron $j$, $\lambda_\mathrm{s}=\exp\left(-\frac{\Delta t}{\tau_\mathrm{s}}\right)$ and $\lambda_\mathrm{m}=\exp\left(-\frac{\Delta t}{\tau_\mathrm{m}}\right)$, where $\Delta t$ is the time step, $\tau_\mathrm{s}$ is the synaptic time constant, and $\tau_\mathrm{m}$ is the membrane time constant.
While the first two equations characterize the linear temporal dynamics of the synaptic current $I$ and membrane potential $U$, the last equation captures the non-linearity of the neuron, its spiking output $S$.
Thus in this formulation, we can think of a \ac{LIF} neuron as a binary Perceptron whose inputs are first processed through a linear filter cascade, i.e., the neuronal dynamics, and additionally have a reset mechanism given by the last factor in Eq.~\ref{eq:potential}.

In the stochastic case, this filter does not change.
In fact, we keep the same equations for synaptic current (Eq.~\eqref{eq:current}) and membrane potential (Eq.~\eqref{eq:potential}).
However, instead of a deterministic Heaviside function as in
Eq.~\eqref{eq:spike_gen}, we use a stochastic spike generation mechanism with escape noise, which is independent for each neuron $i$ given its membrane potential
\begin{eqnarray}
    p_i[n] &=& \sigma_{\beta_\mathrm{N}}(U_i[n]-\theta)
    \label{eq:prob}\\
    S_i[n] &\sim& \mathrm{Ber}(p_i[n])~.
    \label{eq:stoch_spike}
\end{eqnarray}
Again, this mechanism is in direct equivalence with the stochastic Perceptron case (cf. Eq.~\eqref{eq:stoch_Perceptron}) and permissive to computing smoothed stochastic derivatives.

The derivative of the current and the membrane potential with respect to the weights induce their own dynamics:
\begin{eqnarray*}
    \frac{d}{dw_{ij}}I_i[n+1] &=& \lambda_\mathrm{s}\frac{d}{dw_{ij}}I_i[n]+ S_j[n]~, \\
\end{eqnarray*}
\begin{eqnarray}
    \frac{d}{dw_{ij}}U_i[n+1]
    &=& \lambda_\mathrm{m}(1-S_i[n])\frac{d}{dw_{ij}}U_i[n] \nonumber\\
    &&+(1-\lambda_\mathrm{m})(1-S_i[n])\frac{d}{dw_{ij}}I_i[n] \nonumber\\
    &&-\left(\lambda_\mathrm{m}U_i[n] + (1-\lambda_\mathrm{m})I_i[n]\right)\frac{d}{dw_{ij}}S_i[n]~,
    \label{eq:u_diff}
\end{eqnarray}
where we used the product rule to include the derivative of the reset term.
To compute the smoothed stochastic derivative of the spike train, we use the affine combination of the left and right smoothed stochastic derivatives of a Bernoulli random variable according to \citet{arya_automatic_2022} to get
\begin{eqnarray}
    \frac{d}{dw_{ij}}S_i[n] &\leftarrow& \underbrace{\frac{d}{dp_i[n]}\mathrm{Ber}(p_i[n])}_{=1}\frac{d}{dU_i[n]}\sigma_{\beta_\mathrm{N}}(U_i[n])\frac{d}{dw_{ij}}U_i[n] \nonumber \\
    &=& \underbrace{\frac{d}{dU_i[n]}\sigma_{\beta_\mathrm{N}}(U_i[n])}_{\mathrm{SD}}\frac{d}{dw_{ij}}U_i[n] ~.
    \label{eq:s_diff}
\end{eqnarray}
Again, we find that this exactly recovers \acp{SG} as introduced in \citet{zenke_superspike_2018} for $\beta_\mathrm{SG} = \beta_\mathrm{N}$ when conditioning on the deterministic spiking path and, hence, confirms the equality between \acp{SG} and smoothed stochastic derivatives.
Thus, our analysis establishes a link between the escape noise function of stochastic \ac{LIF} neurons and the shape of the \ac{SD} in multi-layer \acp{SNN}.
A similar connection between escape noise and exact derivatives in expectation was previously described in shallow \acp{SNN} \citep{pfister_optimal_2006, brea_matching_2013}.

Going further, we can use Eq.~\eqref{eq:s_diff} to write Eq.~\eqref{eq:u_diff} such that it only depends on current and membrane potential derivatives
\begin{eqnarray}
    \frac{d}{dw_{ij}}U_i[n+1]
    &=& \left(\lambda_\mathrm{m}\left(1-S_i[n]\right) - \underbrace{\sigma'_{\beta_\mathrm{N}}\left(U_i[n]\right)\cdot\left(\lambda_\mathrm{m}U_i[n] + \left(1-\lambda_\mathrm{m}\right)I_i[n]\right)}_{\text{derivative through reset}}\right)\frac{dU_i[n]}{dw_{ij}} \label{eq:diff_reset}  \\
    &&+\left(1-\lambda_\mathrm{m}\right)\left(1-S_i[n]\right)\frac{dI_i[n]}{dw_{ij}} \nonumber
\end{eqnarray}
and can be computed forward in time.
Most \ac{SNN} simulators avoid \ac{BP} through the reset term \citep{zenke_superspike_2018,eshraghian_training_2023} as this empirically improves performance for poorly scaled \acp{SG} \citep{zenke_remarkable_2021}.
Therefore, it is common practice to set the right-hand term in the parenthesis in Eq.~\eqref{eq:diff_reset}, i.e., the derivative through the reset term, to zero.
Conversely, strict adherence to \ac{stochAD} would suggest keeping the reset term when back-propagating.
However, similar to \citet{zenke_remarkable_2021}, we find no difference in performance whether we back-propagate through the reset or not when the \ac{SG} is scaled with $\frac{1}{\beta_\mathrm{SG}}$ (see supplementary Fig.~\ref{sfig:train}), while without this scaling, \ac{BP} through the reset negatively affects performance.
Overall, we have shown that our results obtained from the analysis of Perceptrons are transferable to \acp{SNN} and hence confirm again the equality between \acp{SG} and smoothed stochastic derivatives in the \ac{stochAD} framework.

\begin{figure}[tpb]
    \centering
    \includegraphics[]{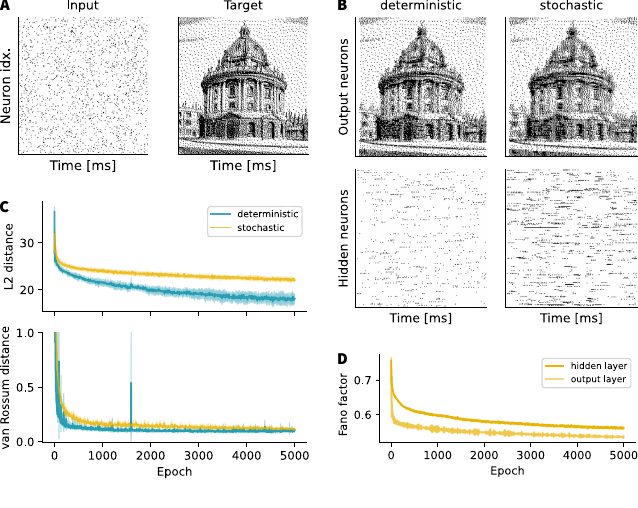}
    \caption{\textbf{\acp{SG} successfully train stochastic \acp{SNN} on a spike train matching task.}
        \textbf{(A)}~Spike raster plots of the input and target spike trains of the spike train matching task.
        Time is shown on the $x$-axis, the neuron indices are on the $y$-axis and each dot is a spike.
        The task is to convert the given frozen Poisson input spike pattern (left) into a structured output spike pattern depicting the Radcliffe Camera in Oxford (right).
        \textbf{(B)}~Top: Output spike raster plots after training of the deterministic (left) and the stochastic \acp{SNN} (right).
        Although both methods faithfully reproduce the overall target structure, the deterministic network is slightly better at matching the exact timing.
        Bottom: Hidden layer spike raster plots. Despite the similar output, the two networks show visibly different hidden layer activity patterns.
        \textbf{(C)}~$L_2$ loss (top) and van Rossum distance (bottom) throughout training for the two network models.
        While the deterministic network outperforms the stochastic one in terms of $L_2$ distance, the difference is negligible for the van Rossum distance.
        \textbf{(D)}~Average Fano factor throughout training in the hidden and output layer of the stochastic network. Although the stochastic network reduces its variability during training to match the deterministic target, its hidden layer still displays substantial variability at the end of training.}
    \label{fig:train_oxford}
\end{figure}

\section{Surrogate gradients are ideally suited for training stochastic \acp{SNN}}
\label{sec: empirical_results}
Above we have seen that \ac{SG} descent is theoretically justified by \ac{stochAD} albeit better supported for stochastic spiking neuron models.
This finding also suggests that \ac{SG} descent is suitable for training stochastic \acp{SNN}, with deterministic \acp{SNN} being a special case (cf.\ Fig.~\ref{fig:existing_frameworks}).
Next, we wanted to confirm this insight numerically.
To that end, we trained deterministic and stochastic \acp{SNN} with \ac{SG} descent on a deterministic spike train matching and a classification task.

For the spike train matching task, we assumed 200 input neurons with frozen Poisson spike trains.
We then set up a feed-forward \ac{SNN} with one hidden layer, initialized in the fluctuation-driven regime \citep{rossbroich_fluctuation-driven_2022}.
We used supervised \ac{SG} training in a deterministic and a stochastic version of the same network to match 200 target spike trains which were given by a dithered picture of the Radcliffe Camera (Fig.~\ref{fig:train_oxford}A; Methods Section~\ref{sec:met:spiketrainmatching}).
Both networks learned the task, albeit with visibly different hidden layer activity (Fig.~\ref{fig:train_oxford}B).
The deterministic version outperformed the stochastic \ac{SNN} in terms of $L_2$-distance at 1~ms, i.e. the temporal resolution of the simulation.
However, when comparing their outputs according to the van Rossum distance \citep{rossum_novel_2001} with an alpha-shaped filter kernel equivalent to the $\epsilon$-kernel of the \ac{LIF} neuron ($\tau_\mathrm{mem}=10~\mathrm{ms}, \tau_\mathrm{syn}=5~\mathrm{ms}$), we found no difference in loss between the stochastic and deterministic networks (Fig.~\ref{fig:train_oxford}C).
Finally,  we observed that the Fano factor, a measure of stochasticity, of the stochastic \ac{SNN} dropped during training to better accommodate the deterministic target (Fig.~\ref{fig:train_oxford}D).
In summary, we find that the stochastic network exhibits comparable performance to the deterministic network.
Thus \acp{SG} are suitable for training stochastic \acp{SNN}.

\begin{figure}[tpb]
    \centering
    \includegraphics[]{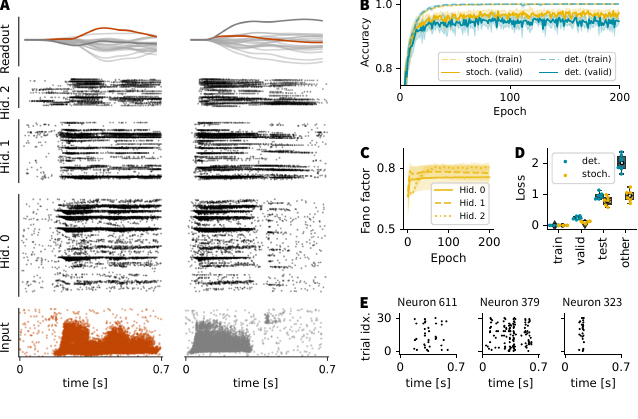}
    \caption{\textbf{\acp{SG} can successfully train stochastic \acp{CSNN}.}
        \textbf{(A)~}Snapshot of network activity for two correctly classified sample inputs from the \ac{SHD} dataset.
        Top: readout unit activity over time, the colored line indicates the activity of the unit that corresponds to the correct class. Below: Spike raster plots of the three convolutional hidden layers.
        The spike raster plots of the inputs are shown at the bottom in red and gray.
        Time is on the x-axis, the neuron index is on the y-axis and each dot represents a spike.
        \textbf{(B)}~Learning curves for training (dashed) and validation (solid) accuracy for the stochastic and the deterministic case (average over $n=6$ trials $\pm$ std).
        \textbf{(C)}~The mean Fano factor of the different layers in the stochastic network throughout training ($\pm$ std, $n=3$).
        \textbf{(D)}~The first three pairs of boxes show train, validation, and test loss of the \ac{CSNN} as in (A) for the stochastic and the deterministic case for $n=6$ random initializations.
        The rightmost boxes show test loss for the opposite activation function.
        This means the network trained deterministically is tested with stochastic activation and vice versa.
        \textbf{(E)}~Raster plots over trials of the spiking activity of three randomly chosen units from the second hidden layer.
        The units show clear trial-to-trial variability reminiscent of cortical activity.
    }
    \label{fig:train_CSNN}
\end{figure}

To verify that this finding generalizes to a more complex task, we trained a \ac{CSNN} with three hidden layers on the \ac{SHD} dataset \citep{cramer_heidelberg_2022} with maximum-over-time readout (Methods Section~\ref{sec:met:classification} and Fig.~\ref{fig:train_CSNN}A).
Like above, the stochastic network learned to solve the task with
comparable training and validation accuracy to the deterministic network (Fig.~\ref{fig:train_CSNN}B).
Specifically, we found that the stochastic \ac{CSNN} achieved a validation accuracy of $97.2\pm 1.0~\%$, (test $85.1\pm 1.3~\%$), compared to $94.7\pm 0.8~\%$ (test $84.3\pm~1.0\%$) for the deterministic \ac{CSNN}.
Furthermore, in the case of the stochastic \ac{SNN}, we left the escape noise active during validation and testing.
However, one can see that the stochastic network performs equally well when tested in a deterministic environment (Fig.~\ref{fig:train_CSNN}D, other).
Conversely, the deterministically trained network does not perform well when evaluated under stochastic spike generation.
In contrast to the previous task, we found that the Fano factor remained high during training (Fig.~\ref{fig:train_CSNN}C), i.e. stochasticity is preserved.
This is also reflected in a high trial-to-trial variability as shown in Fig.~\ref{fig:train_CSNN}E.
One can see that the spiking activity for three example neurons shows a high variability across trials for the same input.
Thus, even for a more complex task, \ac{SG} descent is well suited for training stochastic \acp{SNN}, and stochasticity is preserved after training.

 \section{Discussion}
\label{sec: discussion}

We have analyzed the theoretical foundations of the widely used \ac{SG} descent method for multi-layer \acp{SNN} and shown that \acp{SG} can be derived for stochastic networks from the \ac{stochAD} framework \citep{arya_automatic_2022}.
The \ac{stochAD} framework provides a principled way to use \ac{AD} in discrete stochastic systems, directly applicable to \ac{SG} descent in stochastic \acp{SNN}.
Our analysis is based on smoothed stochastic derivatives introduced within the \ac{stochAD} framework and stipulates that the \ac{SD} should match the derivative of the escape noise function of the stochastic neuron model.
We confirmed in simulations that \ac{SG} descent allows training well-performing stochastic \acp{SNN} with substantial variability comparable to neurobiology.
Our work shows that \acp{SG} enjoy better theoretical support in stochastic \acp{SNN}, which removes some ambiguity of choice of the \ac{SD} and also sheds light on deterministic \acp{SG} as a special case.

One of our key findings is the relation between the functional shape of the escape noise in stochastic neuron models and the \ac{SD}.
This relation is evident in single neurons, where the solution found by \acp{SG} and \ac{stochAD} is equal to the exact derivative of the expected output.
This relation has also been described in previous work on shallow networks \citep{pfister_optimal_2006, brea_matching_2013}, where the derivative of the escape noise directly shows up when computing exact derivatives in expectation.
In this article, we found that the relation extends to multi-layer \acp{SNN} when applying \ac{stochAD} as a theoretical framework.
The link between the \ac{SD} shape and escape noise provides a rigorous mathematical justification for a particular \ac{SD} choice, which in previous work was typically chosen heuristically.

The issue of non-existing derivatives is not unique to \acp{SNN} but is a well-known challenge when dealing with discrete random variables and their derivatives.
It hence arises in various contexts, such as in general binary neural networks \citep{courbariaux_binarized_2016}, sigmoid belief nets \citep{neal_connectionist_1992}, or when dealing with categorical distributions \citep{maddison_concrete_2017,jang_categorical_2016}.
To address this challenge in arbitrary functions, including discrete random variables, the score function estimator, also called REINFORCE \citep{williams_simple_1992}, is often applied, as it provides unbiased estimates.
This estimator can be computed through stochastic computation graphs \citep{schulman_gradient_2015}.
While \citet{arya_automatic_2022} use coupled randomness to reduce variance, there are also methods building on reinforcement learning for low variance gradient estimation \citep{weber_credit_2019}.

Conversely, to address the challenge of \ac{BP} through non-differentiable functions in neural networks instead, a commonly used solution is the \ac{STE} \citep{hinton_lectures_2012,bengio_estimating_2013-1}, which replaces the derivative of a stochastic threshold function with one, also called the identity \ac{STE} \citep{yin_understanding_2018}.
This solution relates closely to \acp{SG} used in \acp{SNN}.
However, unlike the \ac{STE}, \acp{SG} are commonly used in deterministic networks and comprise a nonlinear \ac{SD} like the derivative of a scaled fast sigmoid \citep{neftci_surrogate_2019}.
While both are successful in practice, we still lack a complete theoretical understanding.
Several studies have analyzed the theoretical aspects of the \ac{STE}, looking at the problem from various angles.
However, they have each analyzed different versions of the \ac{STE}, and a complete picture of the underlying theory is still missing.
Although none of them analyzed \acp{STE} in \acp{SNN}, i.e., \acp{SG}, examining their results helps to put \acp{SG} into perspective.
\citet{bengio_estimating_2013-1} originally introduced the identity \ac{STE} as a biased estimator, saying that it guarantees the correct sign only in shallow networks.
\citet{yin_understanding_2018} studied coarse gradients, including the \ac{STE}.
They studied this either in a network with only one nonlinearity or in an activation quantized network with a quantized \ac{ReLU} as opposed to a Heaviside activation function, as we have in the case of \acp{SNN}.
They analyzed training stability and investigated which choice of \acp{STE} leads to useful weight updates.
They concluded that the identity \ac{STE} does not, while the \ac{ReLU} and clipped \ac{ReLU} versions do.
They further found that the identity \ac{STE} might be repelled from certain minima.
\citet{liu_bridging_2023} showed that the identity \ac{STE} provides a first-order approximation to the gradient, which was previously shown by \citet{tokui_evaluating_2017} specifically for Bernoulli random variables.
However, other \acp{STE} were not considered, such as e.g. the sigmoidal \ac{STE}, which would correspond to our \ac{SG}.
Another approach to dealing with discrete distributions was pursued by \citet{jang_categorical_2016,maddison_concrete_2017}. They proposed a solution similar to the reparametrization trick but combined with a smooth relaxation of the categorical distribution.
This relaxation amounts to training a network with continuous representations by sampling from a Gumbel-Softmax.
At the same time, after training, the temperature of the Gumbel-Softmax distribution is set to zero to obtain a discrete output.
Hence, while providing discrete output after training, such a network is trained with continuous neuronal outputs rather than spikes.

Interestingly, most of these solutions deal with discrete functions in the stochastic case, such as stochastic binary Perceptrons or Bernoulli random variables, where noise is used for smoothing.
But as we saw above, probabilistic approaches in the context of \acp{SPM} cannot be applied to multi-layer \acp{SNN} and \ac{BP} without approximations.
In the past, stochastic \ac{SNN} models have been implemented mainly for theoretical and biological plausibility reasons \citep{pfister_optimal_2006,brea_matching_2013}.
For example \citet{brea_matching_2013} showed a link between the escape noise and the voltage-triplet rule \citep{clopath_voltage_2010} in shallow networks.
Furthermore, there were also approaches that enabled the training of multi-layer stochastic \acp{SNN} with \ac{AD}, such as the MultilayerSpiker proposed by \citet{gardner_learning_2015}, which uses the spike train itself as the \ac{SD} of a spike train.
Today, \acp{SNN} are mostly implemented as networks of deterministic \ac{LIF} neurons with a Heaviside function as spiking non-linearity, as they tend to have higher performance.
Recently, however, \citet{ma_exploiting_2023} found empirically that stochastic \acp{SNN} with different types of escape noise can be trained with \ac{SG} descent to high performance.
While they showed a connection to deterministic \ac{SG} descent, they did not discuss the implications for multi-layer networks with stateful \ac{LIF} neurons.
Again, the effectiveness of \ac{SG} descent in stochastic \acp{SNN} was confirmed, as suggested by the connection to the \ac{stochAD} framework found here.
Thus, our work not only provides empirical confirmation of the effectiveness of \acp{SG} in training stochastic \acp{SNN} but also a comprehensive theoretical explanation for their success in multi-layer networks.

To gain analytical insight into the theory behind \acp{SG}, we had to make some simplifying assumptions.
For example, we performed most of our theoretical analysis on Perceptrons which, unlike \ac{LIF} neurons, have no memory.
Nevertheless, our results readily transfer to \ac{LIF} neurons in discrete time, as such networks can be understood as a special type of binary recurrent neural networks \citep{neftci_surrogate_2019} where the output is given by a Heaviside step function.
However, one difference is that, unlike Perceptrons, \ac{LIF} models have a reset after the spike, which should also be considered when calculating gradients.
Smoothed stochastic derivatives allow and suggest back-propagating through the reset mechanism, but previous work has shown that it is beneficial to exclude the reset mechanism from the backward pass, as this leads to more robust performance, especially when the \ac{SG} is not scaled to one \citep{zenke_superspike_2018,zenke_remarkable_2021,eshraghian_training_2023}.
Both options are possible when using \ac{SG}-descent, and we have not noticed much difference in performance when scaling the \acp{SD} by $\frac{1}{\beta_\mathrm{SG}}$ to one.
However, omitting this scaling negatively impacts performance when back-propagation through the reset (see Supplementary Fig. ~\ref{sfig:train}C and compare the asymptotic SuperSpike in \citet{zenke_superspike_2018}).
While omitting this scaling can potentially be counteracted by an optimizer using per-parameter learning rates, the prefactor due to $\beta_\mathrm{SG}$ in the unscaled case can become quite large as it accumulates across layers and thus still affects performance.

Another limitation is that we derived \acp{SG} for stochastic \ac{SNN} models, while they are most commonly used in deterministic networks.
This discrepancy may in fact point at a possible advantage.
Our simulations suggested that stochastic networks generalize equally well or better than their deterministic counterparts.
This difference may be due to stochastic path selection, which leads to lower selection bias compared to the deterministic case.
Moreover, when employing small weight updates, each update follows a different path in the tree of spike patterns, effectively averaging the updates over multiple trials, albeit with slightly different parameter sets.
Such stochastic averaging helps to reduce the bias present in the smoothed stochastic derivatives.
Alternatively, the difference may be due to escape noise acting as an implicit regularizer reminiscent of dropout that promotes better generalization.

Next, while our work motivates a clear link between the \ac{SD} and a stochastic neuron model with escape noise, it does not say which escape noise to use.
This lack may seem like a weakness that merely moves the ambiguity of choice to a different component of the neuron model.
Instead, doing so reduces two heuristic choices to one.
Furthermore, the escape noise function can be constrained using biological data \citep{jolivet_predicting_2006,pozzorini_automated_2015}.

Moreover, the tasks we used to study stochastic \acp{SNN} were limited.
Historically, stochastic \acp{SNN} have been evaluated on tasks that require exact deterministic spike trains at readout such as in the spike train matching task \citep{pfister_optimal_2006, brea_matching_2013, gardner_learning_2015}.
However, despite being performed with stochastic \acp{SNN} such tasks actually punish stochasticity, at least in the readout.
Therefore, stochastic \acp{SNN} respond by becoming more deterministic during training as we have observed when monitoring the Fano factor (cf.\ Fig.~\ref{fig:train_oxford}).
In our setting, neurons have the possibility to change the effective steepness of the escape noise function by increasing their weights, as it is dependent on $\beta_N\cdot|w|$, and thus get more deterministic.
Therefore, we believe that tasks that do not punish stochasticity, such as classification tasks, are much more suitable for evaluating the performance of stochastic \acp{SNN} as they allow training of well-performing \acp{SNN} that still show substantial variability in their responses.

Finally, our analysis of the deterministic \acp{SG} was also subject to some limitations.
In particular, a comparison with the actual gradient, which is either zero or undefined, would not have worked.
Instead, we analyzed the deviation of the \ac{SG} from the actual gradient in differentiable networks that are non-spiking.
Doing so, we have shown that \acp{SG} in deterministic networks with well-defined gradients deviate from the actual gradients.
In the limiting case $\beta_f\to\infty$, these networks are equivalent to Perceptron networks with step activation functions.
Thus, although we could not perform our analysis directly on deterministic \acp{SNN}, we consider the above approach to be a good approximation since it preserves the main properties of \acp{SG}.

The work in this article motivates several exciting directions for future research.
First, we found that although necessary for training, \acp{SG} are generally not gradients of a surrogate loss and thus not guaranteed to find a local minimum.
Specifically, when replacing the non-existent derivative of a Heaviside with a \ac{SD} the resulting \ac{SG} is generally not a gradient.
It seems unlikely that choosing a different escape noise, or \ac{SD} will alter this result.
Thus in situations requiring stronger theoretical guarantees as those bestowed by gradients, an alternative approach would be to start from a surrogate loss function.
While this approach would, by design, yield a proper gradient, it would be more akin to optimizing a smoothed version of the problem, as is the case when using the Gumbel Softmax or the concrete distribution \citep{jang_categorical_2016, maddison_concrete_2017}, rather than defining a surrogate for an ill-behaved gradient.
However, this approach would also come with the known difficulties for \ac{AD}.
It is presently not clear whether or how they can be overcome by future work.

We also saw that \acp{SG} generally decrease the loss but they do not necessarily find a local minimum of the loss.
This discrepancy is due to the deviation from the actual gradient.
Yet, learning in deterministic \acp{SNN} is only possible due to this deviation, because the actual gradient is almost always zero.
In fact, deviations from the actual gradient are often also desirable even in conventional \ac{ANN} training.
Bias and variance of the gradient estimator can help to generalize \citep{ghosh_how_2023}.
For instance, in \acp{ANN} with a rough loss landscape, mollifying, which induces a deviation from the gradient, is successful in practice \citep{gulcehre_mollifying_2016}.
Regularizers or optimizers, which are often used by default, show another advantage of deviations.
For example, many optimizers rely on momentum to speed up training or avoid getting caught in bad local minima.
Thus, an interesting line of future work is to study what constitutes a good local minimum in the case of \acp{SNN}, which types of biases or deviations from the gradient help to find them, and how this relates to \acp{SG}.

Another promising starting point for future work is tied to the \ac{stochAD} framework and questions pertaining to optimization effectiveness.
In this study, we linked \ac{SG}-descent to smoothed stochastic derivatives, which are biased with respect to the non-smoothed version based on the full stochastic triple.
Thus far, the latter can only be computed in forward-mode \ac{AD}, which is impractical for neural network training, as it comes with an increased computational cost.
An intriguing direction for future research is, therefore, to investigate forward mode \ac{AD} with the full stochastic triple to determine whether training improvements due to the reduction in bias would justify the added computational cost.
Furthermore, it would be worth exploring whether this computational cost could be reduced, for instance, by using mixed-mode \ac{AD}, such as in DECOLLE \citep{kaiser_synaptic_2020} or online spatio-temporal learning \citep{bohnstingl_online_2023} (see \citep{zenke_brain-inspired_2021} for a review).

Finally, an exciting direction for future research is to use our theoretically motivated methodology for training stochastic \acp{SNN} to address biological questions.
In neurobiology, many neurons exhibit trial-to-trial variability.
However, the role of this variability is not well understood.
Here, functional stochastic \ac{SNN} models will help to further our understanding of its role in the brain.
Stochasticity may play possible roles in representational drift \citep{micou_representational_2023}, efficient coding \citep{deneve_efficient_2016}, and for biological implementations of latent generative models \citep{hinton_wake-sleep_1995,rombach_high-resolution_2022}.
Thus, it would be interesting to study how learning influences representational changes in plastic functional stochastic \acp{SNN} and whether we can identify specific dynamic signatures that allow drawing conclusions about the underlying learning mechanisms in the brain.

\medskip

In conclusion, \acp{SG} remain a valuable tool for training both deterministic and stochastic \acp{SNN}.
In this article, we saw that \ac{stochAD} provides a theoretical backbone to \ac{SG} learning which naturally extends to stochastic \acp{SNN}.
Further, it suggests that the choice of the \ac{SD} should be matched to the derivative of the escape noise.
Training such stochastic \acp{SNN} is becoming increasingly relevant for applications with noisy data or noisy hardware substrates and is essential for theoretical neuroscience as it opens the door for studying functional \acp{SNN} with biologically realistic levels of trial-to-trial variability.

 \section{Methods}
\label{sec: methods}
Our Perceptron and \ac{SNN} models were written in Python 3.10.4 and extended either on the stork library \citep{rossbroich_fluctuation-driven_2022} which is based on Pytorch \citep{paszke_pytorch_2019}, or Jax \citep{bradbury_jax_2018}.
The code repository can be found at \url{https://github.com/fmi-basel/surrogate-gradient-theory}.
All models were implemented in discrete time.
The following sections provide more details on the different neuron models before providing all the necessary information on the two learning tasks, including the architecture and parameters used for a given task.

\subsection{Neuron models}
\label{sec:neuron_models}

\paragraph{Perceptron.}
As mentioned in Section \ref{sec:single_neurons}, the Perceptron is a simplified version of the \ac{LIF} neuron model, where there is no memory and no reset.
The membrane dynamics $U_i^l$ for Perceptron $i$ in layer $l$ are given for the special case of a single Perceptron by Eq.~\eqref{eq:det_Perceptron_mem_pot} in the theoretical results section, or in general by
\begin{eqnarray*}
    U_i^l &=& \sum_j w_{ij} S_j + b_i~,
\end{eqnarray*}
where $w_{ij}$ are the feed-forward weights, $S_j$ are the binary outputs (spikes) of the previous layer, and $b_i$ is an optional bias term.

\paragraph{Leaky integrate-and-fire neuron.}
We used an \ac{LIF} neuron model with exponential current-based synapses \citep{gerstner_neuronal_2014}.
In discrete time, the membrane potential $U_i^l[n]$ of neuron $i$ in layer $l$ at time step $n$ is given by
\begin{equation}
    U_i^l[n+1] = \left(\lambda_{mem} \cdot U_i^l[n] + (1-\lambda_{mem}) \cdot I_i^l[n]\right) \cdot (1 - S_i^l[n])~,
    \label{eq:membrane_potential}
\end{equation}
where $I_i^l[n]$ is the input current and $S_i^l[n]$ output spike of the neuron itself, which governs reset dynamics.
With multiplicative reset as above, the neuron stays silent for one time step after each spike.
The membrane decay variable $\lambda_{mem} = \exp\left(-\frac{\Delta t}{\tau_\mathrm{mem}}\right)$ is defined through the membrane time constant $\tau_\mathrm{mem}$ and the chosen time step $\Delta t$.
The input current $I_i^l[n]$ is given by
\begin{equation}
    I_i^l[n+1] = \lambda_{syn} \cdot I_i^l[n] + \underbrace{\sum_{j} w_{ij}^l \cdot S_j^{l-1}[n]}_{\mathrm{feedforward}} + \underbrace{\sum_{k} v_{ik}^l \cdot S_k^l[n]}_{\mathrm{recurrent}}~,
    \label{eq:input_current}
\end{equation}
where $w_{ij}$ and $v_{ik}$ are the feedforward and recurrent synaptic weights, respectively, corresponding to the previous layers' spikes $S_j^{l-1}$ and the same layers' spikes $S_k^{l}$, respectively.
The synaptic decay variable is again given as $\lambda_{syn} = \exp\left(-\frac{\Delta t}{\tau_\mathrm{syn}}\right)$, defined through the synaptic time constant $\tau_\mathrm{syn}$.
At the beginning of each minibatch, the initial membrane potential value of all neurons was set to their resting potential $U_i^l[0] = U_{rest} = 0$ and the initial value for the synaptic current was zero as well, $I_i^l[0]=0$.

\subsection{Spike generation}
\label{sec:spike_generation}
The generation of a spike follows the same mechanism in both, the \ac{LIF} and the Perceptron neuron model.
Depending on the membrane potential value, the activation function generates a spike or not.
The following paragraphs highlight the differences between the deterministic and the stochastic cases.

\paragraph{Deterministic spike generation.}
A spike is generated in every time step, in which the membrane potential crosses a threshold $\theta$.
Hence, the Heaviside step function $H(\cdot)$ is used to generate a spike deterministically at time step $n$:
\begin{equation*}
    S_i^l[n] =  H\left(U_i^l[n] - \theta\right)~.
\end{equation*}

\paragraph{Stochastic spike generation.}
A stochastic neuron with escape noise \citep{gerstner_neuronal_2014} may spike, even if the membrane potential is below the threshold, or vice versa not spike, even if the membrane potential is above the threshold.
Therefore, in discrete time, the probability of spiking in time step $n$ is
\begin{equation}
    p_i^l[n] = f(U_i^l[n] - \theta) = \sigma_{\beta_\mathrm{N}}\left(U_i^l[n] - \theta\right)
    \label{eq:spike_prob}
\end{equation}
for each neuron.
We used $\sigma_{\beta_\mathrm{N}}(x) = \frac{1}{1+\exp(-\beta_\mathrm{N} \cdot x)}$, if not stated otherwise.
The hyperparameter $\beta_\mathrm{N}$ defines the steepness of the sigmoid, with $\beta_\mathrm{N} \rightarrow \infty$ leading to deterministic spiking.
Spikes were then drawn from a Bernoulli distribution with probability $p_i^l[n]$, hence
\begin{equation}
    S_i^l[n] \sim \mathrm{Ber}(p_i^l[n])~.
    \label{eq:stoch_spikes}
\end{equation}
The expected value of spiking equals the spike probability, so $\mathbb{E}\left[S_i^l[n]\right] = p_i^l[n]$.

To emphasize the similarity to deterministic spike generation, we can consider an equivalent formulation, where the stochasticity is absorbed into the threshold such that Equations~\eqref{eq:spike_prob} and \eqref{eq:stoch_spikes} are combined into
\begin{equation}
    S_i^l[n] =  H\left(U_i^l[n] - \Theta\right)~.
    \label{eq:stoch_spikes_v2}
\end{equation}
The threshold $\Theta=\theta +\xi$ is defined by its mean $\theta$, i.e. the deterministic threshold, and a zero-mean noise~$\xi$.
For equivalence to the version with the Bernoulli random variable (Eq.~\eqref{eq:stoch_spikes}), the noise $\xi$ is distributed as the $\mathrm{logit}(x)= \log\left(\frac{1}{1-x}\right)$ where $x$ is drawn from a uniform distribution over the interval $[0, 1]$, i.e. $x\sim \mathcal{U}(0,1)$.

\subsection{Differentiable example network}
\label{sec:met:example_net}
The minimal example network (Fig.~\ref{fig:bias_in_sigmoid_networks}A) was simulated using Jax \citep{bradbury_jax_2018}.
To implement the main essence of \ac{SG} descent in a differentiable network, we constructed a network of Perceptrons with sigmoid activation functions (see Eq.~\eqref{eq:example_net}).
The \ac{SG} was implemented by a less steep sigmoid, which was used instead of the actual activation function on the backward pass.

\paragraph{Integrated (surrogate) gradients.}
In general, the integral of the derivative of the loss should equal again the loss $\int \frac{\partial \mathcal{L}}{\partial w} dw = \mathcal{L}$.
The same holds true for the network output.
In Fig.~\ref{fig:bias_in_sigmoid_networks}, we computed this quantity for $w \in [-0.2, 0.2]$, as well as in two dimensions for the parameters $v_1$ and $v_2$.
In Fig.~\ref{fig:SG_are_not_grad_of_loss}, we did not compute this along a line but instead chose to compute this along a closed path in parameter space, i.e. a circle.
To do so, we integrated the \ac{SG} along a circle in a two-dimensional hyperplane which is spanned by two randomly chosen orthonormal vectors $d_1$ and $d_2$ in parameter space.
The position along the circle is defined by an angle $\alpha$ and the circle is parametrized by $a$ and $b$ such that $a=r\cdot \sin(\alpha)$ and $b=r\cdot \cos(\alpha)$ with $r$ the radius of the circle.
Hence, when integrating along the circle, the weights $\theta$ in the network changed according to the angle $\alpha$
\begin{eqnarray*}
    \theta_\alpha = d_1 \cdot a(\alpha)  + d_2 \cdot b(\alpha)~.
\end{eqnarray*}

\subsection{Learning tasks}
We trained \acp{SNN} to evaluate the differences in learning between the stochastic and deterministic \ac{SNN} models trained with \ac{SG}-descent.
For this purpose, we chose a spike train matching and a classification task to cover different output modalities.

\subsubsection{Spike train matching}
\label{sec:met:spiketrainmatching}

If an \ac{SNN} can learn a precise timing of spikes, it must be able to match its output spike times with some target output spike times.
Therefore in the spike train matching task, we ask, whether both, the stochastic and the deterministic network can learn deterministic output spike trains.
For the training, no optimizer is used and since we use a minibatch size of one, this means we apply batch gradient descent.
The details about the used architecture, neuronal, and training parameters are given in the following paragraphs as well as in Tables~\ref{tab:stoch_params} and \ref{tab:arch_params} for both, the stochastic and the deterministic versions of the network.
The code was written in Jax.

\paragraph{Task.}
The target spike trains to match were generated from a picture of the Radcliffe Camera in Oxford using dithering to create a binary image.
We set this image as our target spike raster, where the x-axis corresponds to time steps and the y-axis is the neuron index.
Hence, the image provides a binary spike train for each readout neuron.
As an input, we used frozen Poisson-distributed spike trains with a per-neuron firing rate of 50 Hz.
This created a one-image dataset that requires 200 input neurons, and 200 output neurons and has a duration of 198 time steps, i.e. 198 ms when using $\Delta t=1~\mathrm{ms}$.

\paragraph{Network architecture.}
The above-described spike train matching task is an easy task, that could also be solved by an \ac{SNN} without a hidden layer.
However, since we knew that in the shallow case without a hidden layer, the only difference between our stochastic and our deterministic versions would lie in the loss, we decided to use a network with one hidden layer.
Hence the used network architecture was a feed-forward network with 200 input units, 200 hidden units, and 200 readout units, run with $\Delta t=1~\mathrm{ms}$ for a total of 198 time steps (see also Table~\ref{tab:arch_params}).
Weights were initialized in the fluctuation-driven regime \citep{rossbroich_fluctuation-driven_2022} with a mean of zero and target membrane fluctuations $\sigma_U=1$.
For the stochastic networks, we averaged the update over ten trials, i.e., we sampled ten paths in the tree of spike patterns, used each of them to compute an \ac{SG}, and took their average to perform the weight update before sampling another path for the next weight update (see Fig.~\ref{fig:existing_frameworks}).

\paragraph{Reset at same time step.}
Matching a target binary image without further constraints with a spike train might require a neuron to spike in two adjacent time steps.
However, this is not possible with the membrane dynamics as in Eq.~\eqref{eq:membrane_potential}, where after every spike the neuron stays silent for one time step.
Hence for this task, we used slightly modified membrane dynamics
\begin{equation*}
    U_i^l[n+1] = \lambda_{mem} \cdot U_i^l[n] \cdot (1 - S_i^l[n]) + (1-\lambda_{mem}) \cdot I_i^l[n]~,
\end{equation*}
where the reset was applied at the same time step as the spike and thus high enough input in the next time step could make the neuron fire again.

\paragraph{Loss functions.}
We trained the network using an $L_2$ loss function
\begin{equation*}
    L_2 = \frac{1}{N} \sum_{i=1}^{M^L} \sum_{n=1}^T \left(S_i^L[n] - \hat{S}_i[n]\right)^2
\end{equation*}
to compute the distance between target spike train $\hat{S}_i$ for readout neuron $i$ and output spike train $S_i^L[n]$ at the readout layer $L$,
where $M^L$ is the number of readout neurons and $T$ is the number of time steps.
Furthermore, we also monitor the van Rossum distance \citep{rossum_novel_2001}
\begin{equation*}
    L_{vR} = \frac{1}{2} \int_{-\infty}^t \left(\left( \alpha \hat S_i - \alpha S_i\right)(s)\right)^2 ds
\end{equation*}
between the target spike train and the output spike train, which might be a better distance for spike trains, since it also takes temporal structure into account by punishing a spike that is five time steps off more than a spike that is only one time step off.
It does so, by convolving the output and target spike train first with a temporal kernel $\alpha$, before computing the squared distance.
We chose $\alpha$ to be equivalent to the $\epsilon$-kernel of the \ac{LIF} neuron
\begin{eqnarray*}
    \alpha(t) = \frac{1}{1-\frac{\tau_\mathrm{mem}}{\tau_\mathrm{syn}}}\left(\exp\left(-\frac{t}{\tau_\mathrm{syn}}\right)-\exp\left(-\frac{t}{\tau_\mathrm{mem}}\right)\right)
\end{eqnarray*}
with $\tau_\mathrm{mem} = 10~\mathrm{ms}, \tau_\mathrm{syn} = 5~\mathrm{ms}$.

\paragraph{Fano factor.}
The Fano factor $F$ was calculated on the hidden layer and output spike trains $S$ as $F = \frac{\sigma^2_S}{\mu_S}$.

\subsubsection{Classification on \ac{SHD}}
\label{sec:met:classification}
To evaluate performance differences, we also evaluated both the stochastic and the deterministic version on a classification task using a multi-layer recurrent \ac{CSNN} (as in \citet{rossbroich_fluctuation-driven_2022}), hence also increasing the difficulty for the stochastic model, which now had to cope with multiple noisy layers.
The details about the used architecture, neuronal, and training parameters are given in the following paragraphs as well as in Tables~\ref{tab:stoch_params} and \ref{tab:arch_params} for both, the stochastic and the deterministic versions of the network.

\paragraph{Task.}
For the classification task, we used the \ac{SHD} dataset \citep{cramer_heidelberg_2022}, which is a real-world auditory dataset containing spoken digits in German and English from different speakers, hence it has 20 classes.
The dataset can be downloaded from \url{https://ieee-dataport.org/open-access/heidelberg-spiking-datasets}.
\citet{cramer_heidelberg_2022} preprocessed the recordings using a biologically inspired cochlea model to create input spike trains for $n=700$ neurons.
The duration of the samples varies, hence we decided to consider only the first $T_\mathrm{SHD} = 700~\mathrm{ms}$ of each sample, which covers $>98\%$ of the original spikes.
For our numerical simulations, we binned the spike trains using a $\Delta t=2~\mathrm{ms}$ and hence we had $\frac{\Delta t}{T_\mathrm{SHD}} = 350$ time steps in this task.
10\% of the training data was used as a validation set, and for testing we used the officially provided test set, which contains only speakers that did not appear in the training set.

\paragraph{Network architecture.}
We used a multi-layer recurrent \ac{CSNN} for the \ac{SHD} classification task (see also Table~\ref{tab:arch_params}).
There were 700 input neurons, that directly get fed the input spikes from the \ac{SHD} dataset.
The network had three recurrently connected hidden layers and used one-dimensional convolution kernels.
Weights were initialized in the fluctuation-driven regime with a mean of zero, target membrane fluctuations $\sigma_U=1$, and the proportion of fluctuations due to the feed-forward input was $\alpha=0.9$.
The network size as well as the parameters for the feed-forward and recurrent convolutional operations are summarized in Table~\ref{tab:arch_params}.
We performed only one trial per update for the stochastic case.

\paragraph{Readout units.}
As opposed to the previous task, a classification task requires special readout units.
The readout units did have the same membrane and current dynamics as a normal \ac{LIF} neuron (Eqns.~\eqref{eq:membrane_potential} and \eqref{eq:input_current}), but they did not spike.
Furthermore, we used a different membrane time constant $\tau_\mathrm{mem}^\mathrm{RO} = T_\mathrm{SHD} = 700~\mathrm{ms}$ for the readout units.
This allowed us to read out their membrane potential for classification (see paragraph on the loss function).

\paragraph{Activity regularization.}
To prevent tonic firing, we used activity regularization to constrain the upper bound of firing activity.
To this end, we constructed an additional loss term as a soft upper bound on the average firing activity of each feature in our convolutional layers.
Hence for every layer, we computed a term
\begin{equation}
    g_{upper}^{l, k} = \left(\left[\frac{1}{M^l}\sum_i^{M^l} \zeta_i^{l, k}-\vartheta_{upper}\right]_+\right)^2~,
\end{equation}
where $M^l = n_{features}\times n_{channels}$ is the number of neurons in layer $l$ and $\zeta_i^{l, k} = \sum_n^T S_i^{l,k}[n]$ is the spike count of neuron $i$ in layer $l$ given input $k$.
We chose the parameter $\vartheta_{upper} = 7$ to constrain the average firing rate to 10 Hz.
Hence, the total upper bound activity regularization loss is given by
\begin{equation}
    L_{upper} = - \lambda_{upper}  \sum_{l=1}^L \sum_{f=1}^{F^l} g_{upper}^{l, k}~,
\end{equation}
where $\lambda_{upper} = 0.01$ was the regularization strength. Those parameters can also be found in Table~\ref{tab:arch_params}.

\paragraph{Loss function.}
As this is a classification task, we used a maximum-over-time readout.
To that end, we used a standard cross-entropy loss
\begin{equation}
    L_{CE} = - \frac{1}{K} \sum_{i=1}^{k} \sum_{c=1}^C y_c^k \log\left(p_c^k\right)
    \label{eq:ce_loss}
\end{equation}
to sum over all samples $K$ and all classes $C$.
The correct class is encoded in $y_c^k$ as a one-hot encoded target for the input $k$.
To compute the single probabilities $p_c^k$ for each class $c$, we first read out the maximum membrane potential values of each readout neuron over simulation time to get the activities
\begin{equation}
    a_c^k = \max_n(U_c^L[n])~.
\end{equation}
From those activities, we computed the probabilities using a Softmax function $p_c^k = \frac{\exp(a_c^k)}{\sum_{c'}^C \exp(a_{c'}^k)}$.

\paragraph{Optimizer.}
We used the \ac{SMORMS3} optimizer \citep{funk_rmsprop_2015, rossbroich_fluctuation-driven_2022}, which chooses a learning rate based on how noisy the gradient is.
The \ac{SMORMS3} optimizer keeps track of the three values $g$, $g_2$ and $m$, which were initialized to $g=g_2 = 0$ and $m=1$. They are updated after every minibatch as follows:

\begin{eqnarray*}
    r &=& \frac{1}{m+1}\\
    g &=& (1-r) \cdot g + r \cdot \left(\frac{\partial\mathcal{L}}{\partial\theta}\right)\\
    g_2 &=& (1-r) \cdot g_2 + r \cdot \left(\frac{\partial\mathcal{L}}{\partial\theta}\right)^2\\
    m  &=& 1 + m \cdot\frac{1 - g^2}{g_2 + \epsilon}~.
\end{eqnarray*}

Given this, \ac{SMORMS3} computes the current effective learning rate as $\eta_\mathrm{current} = \frac{\min\left(\eta, \frac{g^2}{g_2 + \epsilon}\right)}{\sqrt{g_2}+\epsilon}$, where $\eta$ is the initially chosen learning rate and $\epsilon$ is a small constant to avoid division by zero.
Therefore the parameter update is $\Delta \theta = -\eta_\mathrm{current} \cdot \frac{\partial\mathcal{L}}{\partial \theta}$.

\paragraph{Fano factor.}
The Fano factor was calculated after taking a moving average over the spike train $S$ with a window of 10 timesteps (20 ms) to get $S_\mathrm{bin}$.
Subsequently, the Fano factor $F$ was computed as
\begin{eqnarray*}
    F = \frac{\sigma^2_{S_\mathrm{bin}}}{\mu_{S_\mathrm{bin}}}~.
\end{eqnarray*}

\begin{table}[htpb]
    \centering
    \def\arraystretch{1.4}
    \setlength{\tabcolsep}{5pt}
    \captionsetup{labelfont=bf}
    \caption{\textbf{Noise and surrogate function parameters.} Parameters used in our numerical simulations for feed-forward networks on the spike-train matching task and multi-layer recurrent \acp{CSNN} on the \ac{SHD} classification task. We selected the learning rate based on the best validation accuracy (cf. Supplementary Fig.~\ref{sfig:train}B).
        The SuperSpike non-linearity $h(x)$ is the derivative of a fast sigmoid scaled by $\frac{1}{\beta}$: $h(x)=\frac{1}{(\beta|x|+1)^2}$. }

    \begin{tabularx}{\textwidth}{@{\extracolsep{\fill}}p{3.5cm}YYcYY@{}}
        \toprule
                                    & \multicolumn{2}{c}{\textbf{Spike train matching task}}  &                                                         & \multicolumn{2}{c}{\textbf{Classification task}}                                   \\
        \cmidrule{2-3}                                                                                   \cmidrule{5-6}
                                    & stochastic                                              & deterministic                                           &                                                  & stochastic      & deterministic \\
        \midrule
        Number of trials            & 10                                                      & 1                                                       &                                                  & 1               & 1             \\
        Learning rate               & $\eta_\mathrm{hid}= 10^{-05}$ $\eta_\mathrm{out}= 10^{-05}$ & $\eta_\mathrm{hid}= 10^{-05}$ $\eta_\mathrm{out}= 10^{-04}$ &                                                  & 0.01            & 0.01          \\
        \midrule
        \textit{Escape noise}       &                                                         &                                                         &                                                  &                 &               \\
        Function                    & $\sigma(\cdot)$                                         & step                                                    &                                                  & $\sigma(\cdot)$ & step          \\
        Parameter                   & $\beta_\mathrm{hid} = 10$ $\beta_\mathrm{out} = 100$        & -                                                       &                                                  & $\beta = 10$    & -             \\
        \midrule
        \textit{Surrogate gradient} &                                                         &                                                         &                                                  &                 &               \\
        Function                    & $\sigma'(\cdot)$                                        & SuperSpike                                              &                                                  & SuperSpike      & SuperSpike    \\
        Parameter                   & $\beta_\mathrm{hid} = 10$                                 & $\beta = 10$                                            &                                                  & $\beta = 10$    & $\beta = 10$  \\
        \bottomrule
    \end{tabularx}
    \label{tab:stoch_params}
\end{table}

\begin{table}[htpb]
    \centering
    \def\arraystretch{1.4}
    \setlength{\tabcolsep}{5pt}
    \begin{minipage}{\textwidth}
        \captionsetup{labelfont=bf}
        \caption{\textbf{Network and training parameters.} Parameters used in numerical simulations for feed-forward \acp{SNN} on the spike-train matching task and multi-layer recurrent \acp{CSNN} on the \ac{SHD} classification task.}

        \begin{tabular*}{\textwidth}{@{\extracolsep{\fill}}lcc}
            \toprule
            & \textbf{Spike train matching task}         & \textbf{Classification task}                \\
            \midrule
            Dataset                 & Radcliffe Camera                       & \ac{SHD}                               \\
            No. input neurons       & 200                                   & 700                                    \\
            No. hidden neurons      & 200                                   & 16-32-64                               \\
            No. output neurons      & 200                                   & 20                                     \\
            No. training epochs     & 5000                                  & 200                                    \\
            Time step               & 1 ms                                  & 2 ms                                   \\
            Duration                & 198 ms                                & 700 ms                                 \\
            Mini-batch size         & 1                                     & 400                                    \\
            Kernel size (ff)        & -                                     & 21-7-7                                 \\
            Stride (ff)             & -                                     & 10-3-3                                 \\
            Padding (ff)            & -                                     & 0-0-0                                  \\
            Kernel size (rec)       & -                                     & 5                                      \\
            Stride (rec)            & -                                     & 1                                      \\
            Padding (rec)           & -                                     & 2                                      \\
            \midrule
            Loss                                    & $L_2$ or van Rossum distance \citep{rossum_novel_2001}    &  Maximum over time                      \\
            Optimizer                               & None (gradient descent)                                  &  \ac{SMORMS3}\citep{funk_rmsprop_2015}                           \\
            \midrule
            \textit{Neuronal parameters}            &                                                          &                                         \\
            Spike threshold                         & 1                                                        &  1                                      \\
            Resting potential                       & 0                                                        &  0                                      \\
            $\tau_\mathrm{mem}$                            & $10~\mathrm{ms}$                                           &  $20~\mathrm{ms}$                         \\
            $\tau_\mathrm{syn}$                            & $5~\mathrm{ms}$                                            &  $10~\mathrm{ms}$                         \\
            $\tau_\mathrm{mem}^\mathrm{RO}$                       & -                                                        &  $700~\mathrm{ms}$                        \\
            Reset                                   & at same time step                                        &  at next time step                      \\
            \midrule
            \textit{Activity regularizer}           &                                                          &                                         \\
            $\vartheta_\mathrm{upper}$                     & -                                                        &  7                                      \\
            $\lambda_\mathrm{upper}$                       & -                                                        &  0.01                                   \\
            \bottomrule
        \end{tabular*}
        \label{tab:arch_params}
    \end{minipage}

\end{table}

\section*{Acknowledgments}
The authors thank Frank Schäfer, Guillaume Bellec, and Wulfram Gerstner for stimulating discussions.
This work was supported by the Swiss National Science Foundation (Grant Number PCEFP3\_202981) and the Novartis Research Foundation.

\printbibliography

\clearpage
\appendix

\renewcommand{\thefigure}{S\arabic{figure}}
\setcounter{figure}{0}

\renewcommand{\thetable}{S\arabic{table}}
\setcounter{table}{0}

\renewcommand{\thesubsection}{S\arabic{subsection}}

\section{Supplementary figures}
\label{sec:sfigs}

\begin{figure}[htbp]
    \centering
    \includegraphics{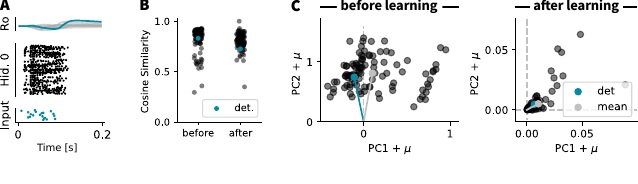}
    \caption{\textbf{The bias induced in \ac{SNN} training due to deterministic \acp{SG} is small.}
        \textbf{(A)}~Spike raster plot of a network trained on the \ac{Randman} dataset \citep{zenke_remarkable_2021}; top: membrane potential of the readout units, middle: spike raster plot of the 128 hidden units, bottom: input data. Time is on the x-axis, the y-axis is the neuron index and each dot is a spike.
        \textbf{(B)}~Cosine similarity of gradients obtained in 100 single trials in a stochastic \ac{SNN} performing the \ac{Randman} task in (A) with respect to the mean gradient over 100 trials before and after learning. Teal shows the cosine similarity between the deterministic and the mean stochastic gradient.
        \textbf{(C)}~First two principal components of the gradients (with the mean added back) obtained in different trials with the stochastic network before (left) and after (right) training on the \ac{Randman} task for 200 epochs. Teal is the deterministic, and silver is the mean gradient.
        One can see that the direction chosen by the deterministic update slightly deviates from the average direction of an update in a stochastic network.
    }
    \label{sfig:bias}
\end{figure}

\begin{figure}[htbp]
    \centering
    \includegraphics[]{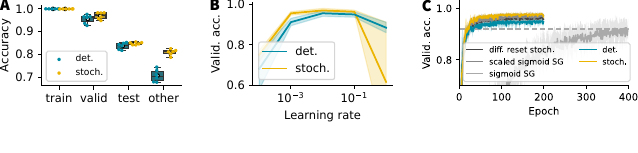}
    \caption{\textbf{Supplementary metrics on the \ac{CSNN} experiments.}
        \textbf{(A)}~Train, validation and test accuracy for the three-layer \ac{CSNN} trained on \ac{SHD} for the stochastic and deterministic case. In the case of the stochastic \ac{SNN}, escape noise is applied during, training, validation, and testing.
        The box labeled "other" shows the performance, if the stochastic network is evaluated on the test set without any escape noise being present and vice versa for the deterministic.
        \textbf{(B)}~Validation accuracy for the stochastic and deterministic networks after training for different learning rates.
        \textbf{(C)}~Validation accuracy for different setups, that applied specific changes to the setup from the main text (compare Fig.~\ref{fig:train_CSNN}): \textit{diff. reset stoch.}: stochastic network trained with \ac{BP} also through the reset term. \textit{Scaled sigmoid \ac{SG}}: stochastic network trained with sigmoid instead of fast sigmoid \ac{SG}, but scaled by $\frac{1}{\beta_{SG}}$, as done in \citet{zenke_superspike_2018} for the fast sigmoid. \textit{Sigmoid \ac{SG}}: same as (B), but without scaling and $\beta_{SG}=1$.
    }
    \label{sfig:train}
\end{figure}

\section{Example: stochastic derivative of a Perceptron as in \ac{stochAD} \citep{arya_automatic_2022}}
\label{sec:example_stochAD_Perceptron}
Let us consider a stochastic binary Perceptron as in Eq.~\eqref{eq:stoch_Perceptron}.
Let us first consider only the derivative of the Bernoulli with respect to the probability of firing $p$, namely $\frac{dy}{dp} = \frac{d}{dp}\mathrm{Ber}(p)$, as we can later apply the chain rule after smoothing.
For the right stochastic derivative (which takes into account jumps from $0$ to $1$), we assume $\epsilon>0$, thus the differential $dy(\epsilon) = y(p+\epsilon) - y(p)$ will be
\begin{equation*}
    dy(\epsilon)(\omega) = \begin{cases}1 & \text{if } 1-p- \epsilon \le \omega < 1-p\\ 0 & \text{otherwise}\end{cases}
\end{equation*}
where $\omega$ is the actual sample drawn from the Bernoulli and we use this fixed randomness to compute the differential.
Given a sample $y(p)(\omega)=1$, there can be no jump, but if we start with $y(p)(\omega)=0$, there is a probability of $\frac{1}{1-p}$ that the output of $y(p+\epsilon)(\omega)$ will jump from zero to one.
A stochastic derivative is written as triple, where the first number is the "almost sure" part $\delta$ of the derivative, the second is the weight (probability) $w$ of a finite jump in the derivative, and finally, we have the alternate value $Y$, which is the new value of the output if the jump occurred.
Hence, in our case, the correct stochastic derivative would be
\begin{equation*}
    (\delta_R, w_R, Y_R) = \begin{cases}(0, \frac{1}{1-p}, 1) & \mathrm{if }~ y(p)(\omega) = 0\\ (0,0,0) & \mathrm{if }~ y(p)(\omega)=1\end{cases}~.
\end{equation*}
For the left stochastic derivative, we would only consider jumps from one to zero.
So the left stochastic derivative is
\begin{equation*}
    (\delta_L, w_L, Y_L) = \begin{cases}(0, 0, 0) & \mathrm{if }~ y(p)(\omega) = 0\\ (0,-\frac{1}{p},0) & \mathrm{if }~y(p)(\omega)=1\end{cases}~.
\end{equation*}
To use stochastic derivatives with \ac{BP}, one needs to smooth them first.
This, however, is no longer an unbiased solution.
The smoothed stochastic derivative is defined as $\widetilde{\delta} = \mathbb{E}\left[\delta + w(Y-y(p))|y(p)\right]$ (see also Eq.~\eqref{eq:smoothed_ssd}).
Hence in our case, we have
$\widetilde{\delta}_R = \frac{1}{1-p} \cdot \mathbf{1}_{y(p)=0}$ and $\widetilde{\delta}_L = \frac{1}{p} \cdot \mathbf{1}_{y(p)=1}$.
We know from~\eqref{eq:stoch_Perceptron}, that $p=\sigma_\beta(u)$ and we can compute the continuous part of the derivative, e.g. $\frac{d}{du}p = \beta\sigma_\beta(u)\cdot(1-\sigma_\beta(u))$ by applying the chain rule.
Put together, we end up with
\begin{eqnarray*}
    \widetilde{\delta}_R^{tot} &=& \frac{1}{1-p} \cdot \mathbf{1}_{y(p)=0} \cdot \frac{dp}{du}\\
    &=& \frac{1}{1-\sigma_\beta(u)} \cdot \mathbf{1}_{y(p)=0}\cdot \beta \sigma_\beta(u)(1-\sigma_\beta(u))\\
    &=& \beta \sigma_\beta(u) \cdot \mathbf{1}_{y(p)=0}
\end{eqnarray*}
for the smoothed right stochastic derivative, and
\begin{eqnarray*}
    \widetilde{\delta}_L^{tot} &=& \frac{1}{p} \cdot \mathbf{1}_{y(p)=1} \cdot \frac{dp}{du}\\
    &=& \frac{1}{\sigma_\beta(u)} \cdot \mathbf{1}_{y(p)=1}\cdot \beta \sigma_\beta(u)(1-\sigma_\beta(u))\\
    &=& \beta (1-\sigma_\beta(u)) \cdot \mathbf{1}_{y(p)=1}
\end{eqnarray*}
for the smoothed left stochastic derivative.
\begin{figure}[htb]
    \centering
    \includegraphics[width=0.3\textwidth]{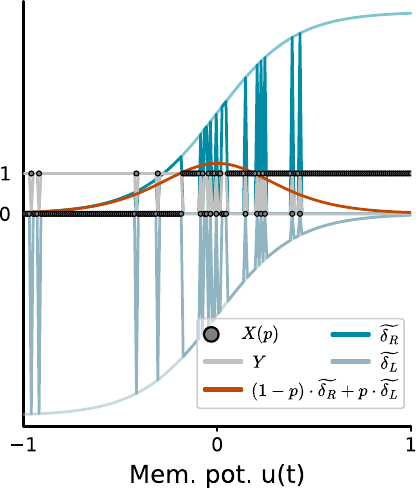}
    \caption{\textbf{Smoothed stochastic derivatives:} Any affine combination of a smoothed left (light blue) and a smoothed right (teal) stochastic derivative is a valid stochastic derivative (red) given a specific realization $X(p)$.}
    \label{sfig:stochAD}
\end{figure}
Now since every affine combination of the left and the right smoothed stochastic derivative is a valid smoothed stochastic derivative, we can choose our smoothed stochastic derivative to be $(1-p) \cdot \widetilde{\delta}_R^{tot} + p\cdot\widetilde{\delta}_L^{tot}$ which evaluates to
\begin{eqnarray*}
    \widetilde{\delta}^{tot} &=& (1-\sigma_\beta(u))  \cdot \beta \sigma_\beta(u) \cdot \mathbf{1}_{y(p)=0} + \sigma_\beta(u) \cdot \beta (1-\sigma_\beta(u)) \cdot \mathbf{1}_{y(p)=1}\\
    &=& \beta\cdot\sigma_\beta(u) \cdot (1-\sigma_\beta(u))
\end{eqnarray*}
Therefore, when using a specific affine combination of the left and the right smoothed stochastic derivatives, we can write $\frac{dy}{du} = \beta\cdot\sigma_\beta(u) \cdot (1-\sigma_\beta(u)) = \beta \cdot\sigma_\beta'(u)$ thereby exactly recovering \acp{SD} in a stochastic network.
Furthermore, we find the same expression for the \ac{SD} independent of the outcome of the Bernoulli random variable.
The \ac{SD} and the smoothed left and right stochastic derivatives for one sample of a stochastic Perceptron are shown in Fig.~\ref{sfig:stochAD}.

\end{document}